\documentclass[10pt]{article} % For LaTeX2e
\usepackage[preprint]{tmlr}

\usepackage{amsmath,amsfonts,bm}

\def\eqref#1{equation~\ref{#1}}
\def\1{\bm{1}}

\DeclareMathAlphabet{\mathsfit}{\encodingdefault}{\sfdefault}{m}{sl}
\SetMathAlphabet{\mathsfit}{bold}{\encodingdefault}{\sfdefault}{bx}{n}

\DeclareMathOperator*{\argmax}{arg\,max}
\DeclareMathOperator*{\argmin}{arg\,min}

\usepackage{hyperref}
\usepackage{url}
\usepackage{algorithm}
\usepackage{algcompatible}
\usepackage{booktabs}
 \usepackage{multirow}
 \usepackage{subcaption}

\title{PICore: Physics-Informed Unsupervised Coreset Selection for Data Efficient Neural Operator Training}
\usepackage{amsmath}
\usepackage{graphicx}

\author{\name Anirudh Satheesh \email anirudhs@terpmail.umd.edu \\
\addr Department of Computer Science\\
University of Maryland, College Park
\AND
\name Anant Khandelwal \email akhandelwal79@gatech.edu \\
\addr Georgia Institute of Technology
\AND
\name Mucong Ding \email mcding@umd.edu\\
\addr Department of Computer Science \\ University of Maryland, College Park
\AND
\name Radu Balan \email rvbalan@umd.edu \\
\addr Department of Mathematics \\ Center for Scientific Computation and Mathematical Modeling  \\ University of Maryland, College Park}

\def\month{MM}  % Insert correct month for camera-ready version
\def\year{YYYY} % Insert correct year for camera-ready version
\def\openreview{\url{https://openreview.net/forum?id=XXXX}} % Insert correct link to OpenReview for camera-ready version

\begin{document}
\newcommand{\rebuttal}[1]{\textcolor{black}{#1}}

\maketitle

\begin{abstract}
Neural operators offer a powerful paradigm for solving partial differential equations (PDEs) that cannot be solved analytically by learning mappings between function spaces. However, there are two main bottlenecks in training neural operators: they require a significant amount of training data to learn these mappings, and this data needs to be labeled, which can only be accessed via expensive simulations with numerical solvers. To alleviate both of these issues simultaneously, we propose PICore, an unsupervised coreset selection framework that identifies the most informative training samples without requiring access to ground-truth PDE solutions. PICore leverages a physics-informed loss to select unlabeled inputs by their potential contribution to operator learning. After selecting a compact subset of inputs, only those samples are simulated using numerical solvers to generate labels, reducing annotation costs. We then train the neural operator on the reduced labeled dataset, significantly decreasing training time as well. Across four diverse PDE benchmarks and multiple coreset selection strategies, PICore achieves up to \(78\%\) average increase in training  relative to supervised coreset selection methods with minimal changes in accuracy.
\end{abstract}

\section{Introduction}
Partial differential equations (PDEs) are foundational to modeling complex physical systems across science and engineering, from fluid dynamics to quantum mechanics. Most PDEs are non-analytic and need to be solved numerically via Finite Difference Methods (FDMs), Finite Element Methods (FEMs), and Finite Volume Methods (FVMs) \cite{cyrus1968accuracy, Johnson1988-em, eriksson_fem, LeVeque_2002}. However, while these approaches yield high accuracy, they are computationally expensive because they require a simulation to be run to obtain a solution. This is especially true for high-resolution or multi-resolution PDEs, where simulations need to be re-run for each resolution.

Operator learning has emerged as a tool for accelerating PDE solutions by developing data-driven approximations using neural networks instead of traditional grid-based discretizations. Neural operators \citep{kovachki2023neural} are a family of neural networks that learn mappings between function spaces, such as initial conditions to solutions, which allows for resolution-invariant predictions. Models such as Fourier Neural Operator (FNO) \citep{li2020fourier} and U-Net Neural Operator (UNO) \citep{rahman2023uno} have shown state-of-the-art performance on various PDE benchmarks, and the ability to generalize to higher-order resolutions with minimal performance drops. Additional work, such as Physics Informed Neural Operator (PINO) \citep{li2024physics} and Markov Neural Operator (MNO) \citep{li2021learning}, incorporates additional losses into neural operator training to improve performance and increase convergence speed.

% PINO uses a physics-informed loss to constrain the training process to conform to the underlying PDE, improving generalization to higher resolutions. MNO uses a dissipativity regularizer to ensure the learned solution does not stray too far from the initial condition.

Despite these advantages, there are two main data limitations of neural operators. First, they require significant amounts of training data to learn these mappings. Since PDE solvers require high-resolution data over several time frames for accurate training, such training data can be several gigabytes large \citep{PDEBench2022}. This poses a challenge for training in resource-constrained systems where such models would be trained and deployed, such as for weather prediction \citep{pathak2022fourcastnet, bonev2023spherical} and carbon storage \citep{tang2024multi}. Secondly, this training data needs to be labeled by including both the initial condition and the ground truth solution. While generating initial conditions is cheap, as they can usually be sampled from a prior distribution, generating ground truth data requires running the full simulation through numerical solvers. 

Coreset selection \citep{agarwal2005geometric, sener2017active} is a data-efficient training strategy that identifies a subset of the original training data that is most informative for model learning. Once this subset is identified, training only needs to be done on this subset, significantly reducing training time. However, this requires the full labeled training data to select a subset, which does not alleviate the cost of collecting labels. On the other hand, active learning \citep{9178457, 9234504} minimizes data annotation costs by only labeling a subset of the training data at each iteration. Active learning selects a subset by a proxy metric such as Bayesian \citep{NEURIPS2021_50d2e70c, 8579074} or representation-based methods \citep{YANG2022108836, kim2022defense} at each training iteration, and trains only on that subset. \rebuttal{A limitation of many iterative active learning strategies is that repeatedly alternating between selecting points and updating the model can increase training time and reduce convergence speed}. Thus, we pose the following research question:

\textit{How can we simultaneously reduce training time and labeling ground-truth solutions for Neural Operator learning?}

We address this problem using unsupervised coreset selection by identifying the most informative training samples based on the physics-informed loss \citep{li2024physics}, a criterion that does not require any ground truth labels. \rebuttal{Our approach can also be viewed as a single-shot active learning implementation, where a subset of points is selected in one pass rather than iteratively.} By leveraging this loss, we can prioritize samples likely to improve model performance without the need for expensive simulations. Ground truth labels are then generated only for this selected subset, significantly reducing the overall annotation cost. Finally, we train neural operator models on the reduced, high-quality dataset, leading to faster training times without compromising accuracy.

Our contributions are outlined as follows:
\begin{itemize}
    \item \textbf{We propose PICore, a novel unsupervised framework that uniquely integrates physics-informed losses with coreset selection.} PICore eliminates the need for expensive ground-truth simulations during the data selection phase, simultaneously addressing the data annotation and training bottlenecks in neural operator training.
    \item \textbf{We demonstrate the modularity and generality of the PICore framework.} Our method is not tied to a specific architecture or selection algorithm, and we show its effectiveness across two different neural operators (FNO and UNO) and five distinct coreset selection strategies.
    \item \textbf{We present the first comprehensive benchmark for coreset selection in the context of neural operator learning.} Through extensive experiments on four diverse PDE datasets, we show that PICore achieves competitive accuracy to supervised methods while dramatically improving end-to-end training  by up to 78\% relative to supervised coreset selection.
\end{itemize}

\section{Related Work}
\subsection{Neural Operators}
While typical deep neural nets are used to map and model finite-dimensional vector spaces, such as text embeddings or images, neural operators map infinite-dimensional vector spaces, such as the space of functions \citep{DBLPjournalscorrabs-2108-08481}. Neural operators are then widely used to represent differential equation solutions due to their ability to have a \textit{family} of solutions. In the context of solving partial differential equations, a neural operator can take a function as an input (e.g. temperature at a point) and output a related function (e.g. heat over time at a point).

Among the first modern neural operators, DeepONet \citep{Lu_2021} uses the universal approximation theorem for operators with a branch and trunk network to model inputs and outputs. The Fourier Neural Operator (FNO) \citep{li2020fourier} expands on this by performing kernel operations in Fourier space, which results in a more expressive model with better performance on more challenging PDE datasets, such as Navier Stokes. U-Net Neural Operator (UNO) \citep{rahman2023uno} expands on FNO by using a U-Net based structure to build deeper neural operators, and Convolutional Neural Operator (CNO) \citep{raonic2023convolutional} leverages convolutions to preserve the continuous structure of PDEs, even when discretized. Additional work improves training by incorporating additional losses. Physics Informed Neural Operator (PINO) \citep{li2024physics} uses the physics informed loss to anchor the output to conform to the PDE dynamics, and Markov Neural Operator (MNO) \citep{li2022learning} uses dissipativity regularization to improve accuracy for more chaotic systems. 

\subsection{Data Efficient Machine Learning}
\subsubsection{Coreset Selection}
For problems where training is too expensive or slow, coreset selection can accelerate training while preserving accuracy. Coreset selection methods can be largely categorized into two types: training-free methods that leverage the geometric properties of the data, and training-based methods that use model-specific information to score data points. Training-free methods involve random \citep{guo2022deepcore, gupta2023data} and geometry-informed selection \citep{10.1145/1553374.1553517, chen2012super}. Recent work on training-based methods can be split into three groups: (i) submodular approaches to maximize the coverage of the selected dataset \citep{pmlr-v37-wei15, mirzasoleiman2020coresets, pooladzandi2022adaptive}, (ii) gradient-based approaches to exactly find the influence of a data point \citep{killamsetty2021grad, NEURIPS2021_ac56f8fe}, and (iii) bilevel optimization methods to improve generalization performance \citep{killamsetty2021retrieve, killamsetty2021glister}.

The traditional testbed for coreset selection algorithms has been image classification tasks, but it also has applications in Neural Architecture Search (NAS) \citep{shim2021core}, efficient GAN training \citep{sinha2020small}, continual learning \citep{yoon2021online}, and large language model (LLM) finetuning \citep{zhang2025staff}. However, to the best of our knowledge, coreset selection has not been used for improving the training  of neural operator learning.

\subsubsection{Active Learning}
\rebuttal{In contrast to coreset selection, active learning, over multiple iterations and in an unsupervised environment, chooses previously unannotated data to label and trains on those newly labeled pairs. The key differences are that active learning is unsupervised, choosing training samples with only features and that active learning is done over many iterations instead of in a single shot. Many algorithms transfer from coreset selection to active learning. Aside from the equivalent random selection, there are cluster based methods for active learning to find representative and typical examples \citep{sener2017active, hacohen2022activelearningbudgetopposite} and  uncertainty based methods that find data for which the model is either uncertain or degraded \citep{rahmati2024understanding, houlsby2011bayesian, ash2019deep}. \citet{musekamp2024active} shows several uncertainty based sampling methods and active learning methods for physics informed learning, but these are limited to PINNs and not to neural operators, which are more powerful due to their ability to map between function spaces, but are inherently more difficult to perform active learning.}

\subsection{Data Efficiency for Neural Operators}
The closest existing work to our own is \citet{chen2024data}, which develops an unsupervised pretraining strategy that leverages Masked Autoencoders (MAEs) to learn effective unsupervised representations, which are then used to fine-tune with a smaller ground-truth dataset. However, this indirectly addresses issues with training  and data labeling costs using a two-stage training process, whereas PICore directly addresses both problems in a single training cycle. \citet{hemmasian2024pretraining} avoid running expensive simulations on high-resolution data by pretraining neural operators in low dimensions, but this requires a factorized neural operator such as Factorized Fourier Neural Operator (FFNO) \citep{tran2021factorized}. In contrast, our method is independent of the operator architecture. \citet{li2024multi} uses an active learning strategy to reduce labeling costs from running simulations by maximizing a utility cost ratio. However, this is specific to FNO and only addresses the cost of data annotation and not training efficiency. \rebuttal{Finally, \citep{pmlr-v267-kim25m} uses a surrogate model for active learning, but this only increases efficiency at the timestep level while ours can reduce the data annotation cost of full solutions in a single shot.} 

\section{Preliminaries}
\subsection{Neural Operators for PDE Solution Generation}

Many physical systems can be modeled using partial differential equations (PDEs), which describe the evolution of a function \( u \in \mathcal{U} \) over a domain. A general PDE can be expressed as
\begin{align}  
    \mathcal{F}(u, a) = 0, \quad \text{on } \Omega \subset \mathbb{R}^d,  
\end{align}
where \( a \in \mathcal{A} \) represents input parameters such as boundary conditions, initial conditions, or physical coefficients; \(\mathcal{F}: \mathcal{U} \times \mathcal{A} \to \mathcal{Z}\) is a differentiable and potentially nonlinear operator; and \(\mathcal{A}, \mathcal{U}\) are Banach spaces over the bounded domain \(\Omega\).

For stationary (time-independent) PDEs, the problem takes the form
\begin{equation}
\begin{aligned}
    \mathcal{F}(u, a) &= 0, \quad \text{on } \Omega \subset \mathbb{R}^d, \\
    u &= h, \quad \text{on } \partial\Omega,
\end{aligned}
\end{equation}
where \( h \) defines the boundary condition on the domain boundary \(\partial\Omega\).

For dynamic (time-dependent) PDEs, the input \( a \) is restricted to the initial condition \( u|_{t=0} \), and the operator \(\mathcal{F}\) is defined on the spatiotemporal domain \(\Omega \times \mathcal{T}\):
\begin{equation}
\begin{aligned}
    \mathcal{F}(u, a) &= 0, \quad \text{on } \Omega \times \mathcal{T}, \\
    u &= h, \quad \text{on } \partial\Omega \times \mathcal{T}, \\
    u &= a, \quad \text{on } \Omega \times \{0\},
\end{aligned}
\end{equation}
where \(\mathcal{T} = (0, T)\) denotes the time domain. Examples of both stationary and dynamic PDEs are provided in Section~\ref{sec: PDE Datasets}.

Unlike conventional neural networks that learn pointwise mappings, neural operators approximate solutions by learning mappings between infinite-dimensional function spaces:
\begin{align}
    \mathcal{G}: \mathcal{A} \to \mathcal{U}.
\end{align}
In practice, a PDE dataset consists of pairs \(\{(a_i, u_i)\}_{i=1}^N\), where each \((a_i, u_i)\) corresponds to an input-output solution of the PDE. The neural operator \(\mathcal{G}\) is approximated by \(\mathcal{G}_\theta\) through the optimization
\begin{align}
    \mathcal{G}_\theta = \argmin_{\theta \in \Theta} \frac{1}{N} \sum_{i=1}^N \|\mathcal{G}_\theta(a_i) - u_i\|^2_{L^2(\Omega)},
\end{align}
where \(\Theta\) is a finite-dimensional parameter space.

\subsection{Coreset Selection}
Given a dataset \(D = \{(x_i, y_i)\}_{i=1}^N\), coreset selection aims to find a subset \(S \subseteq D\) such that
\begin{align}
    S = \argmin_{S' \subset D, |S'|=\beta N} \mathbb{E}_{(x_i, y_i) \sim S'} [\mathcal{L}(x_i, y_i; \theta^{S'})]
\end{align}
where \(\beta\) is the percentage of the original dataset selected and \(\theta^{S'}\) is the model trained on \(S\). However, there are \( \left(\begin{array}{c} N \\ \beta N \end{array} \right) = O(2^{NH_2(\beta)})\) possible subsets of size \(\beta N\), so evaluating this objective directly is infeasible for large datasets. Instead, some works leverage a submodular function \(f: 2^{D} \to \mathbb{R}\) which ensures the diminishing return property
\begin{align}
    f(S \cup \{z\}) - f(S) \geq f(T \cup \{z\}) - f(T), \quad \forall S \subseteq T \subseteq D, z \notin T
\end{align}
This results in a greedy selection procedure, significantly reducing the subset search space. Another way to perform coreset selection is to use a scoring function and select the top-$k$ data points. Finally, coreset selection can be represented as a bilevel optimization problem, resulting in the following form
\begin{align}
S &= \argmin_{S' \subset D,\;|S'|=\beta N} \;\; \mathcal{L}\!\bigl(\theta^*(S')\bigr) \quad \text{s.t.} \quad \theta^*(S') = \argmin_{\theta \in \Theta} \sum_{(x_i, y_i)\in S'} \mathcal{L}(x_i, y_i;\theta)
\end{align}

\section{PICore}
\begin{figure}[htbp]
    \centering
    \fbox{\includegraphics[width=\linewidth]{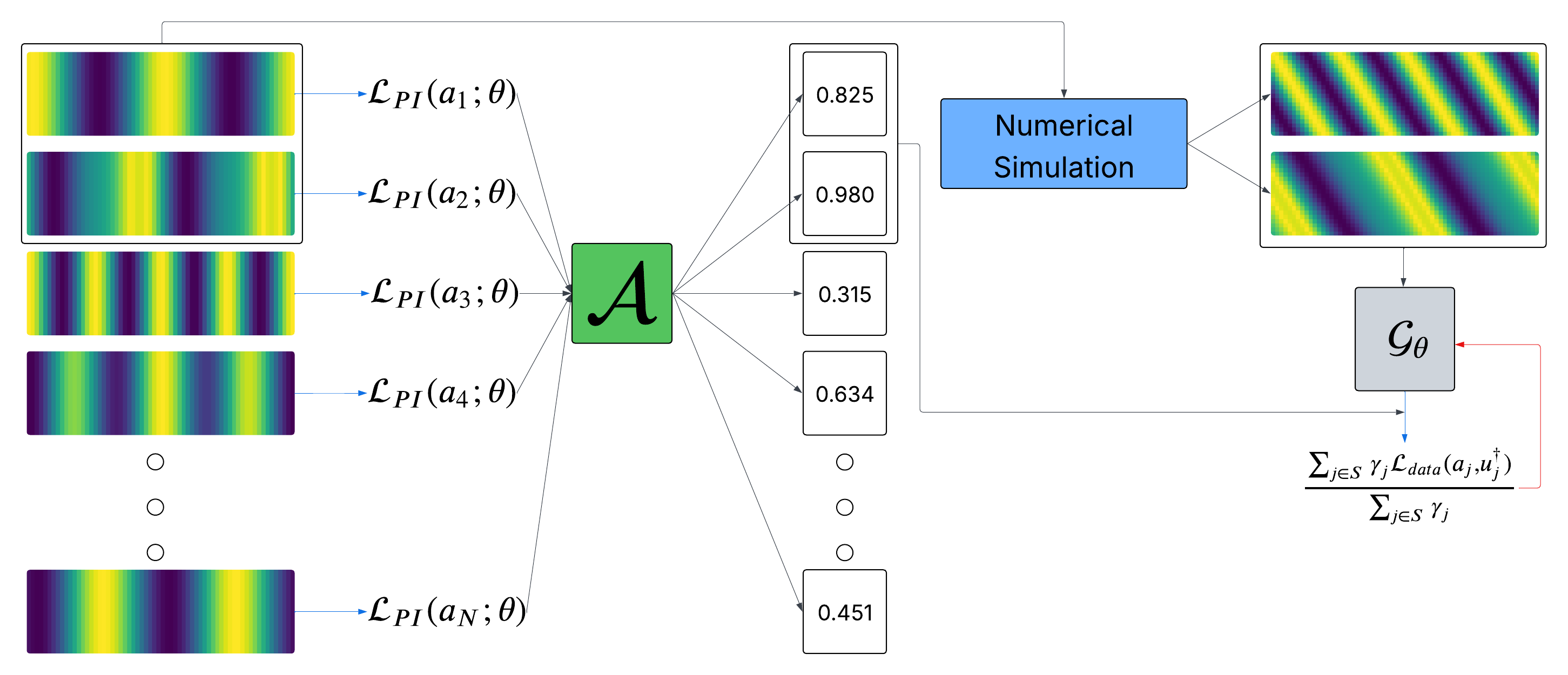}}
    \caption{\textbf{Overview of the PICore Framework.} 
Given a set of initial conditions and a pre-trained (warm-started) neural operator \(\mathcal{G}_\theta\), we compute the physics-informed loss \(\mathcal{L}_{PI}(a_i; \theta)\) for each initial condition \(a_i\). These losses are passed to a coreset selection algorithm \(\mathcal{A}\), which identifies the most informative samples that deviate most from the underlying PDE. Each selected sample is assigned a weight \(\gamma_j\) and is then simulated using a numerical solver to obtain the PDE solutions. The resulting labeled subset is used to update \(\mathcal{G}_\theta\) using a weighted data loss, enabling efficient training by focusing on the most impactful data points. In the figure, blue arrows represent forward passes and red lines represent backward passes respectively.}
    \label{fig:PICore}
\end{figure}

\begin{algorithm}[h]
\caption{PICore: Physics-Informed Coreset Selection for Neural Operators}
\label{alg:picore}
\begin{algorithmic}[1]
\REQUIRE 
Unlabeled dataset $D = \{a_i\}_{i=1}^N$; coreset size $k = \beta N$; learning rate $\alpha$; pretrained operator $\mathcal{G}_\theta$; physics-informed loss $\mathcal{L}_{PI}(a; \theta)$; coreset selection algorithm $\mathcal{A}_{\text{select}}$; warmup steps $T_w$; training steps $T$

\vspace{0.5em}
\STATE \textbf{Warm-start} $\mathcal{G}_\theta$ on unlabeled data using the physics-informed loss:
\FOR{$t = 1$ to $T_w$}
    \FOR{each $a_i \in D$}
        \STATE $\theta \gets \theta - \alpha \nabla_\theta \mathcal{L}_{PI}(a_i; \theta)$
    \ENDFOR
\ENDFOR

\vspace{0.5em}
\STATE \textbf{Score each sample} using physics-informed loss:
\FOR{each $a_i \in D$}
    \STATE $\ell_i \gets \mathcal{L}_{PI}(a_i; \theta)$
\ENDFOR

\vspace{0.5em}
\STATE \textbf{Select coreset indices} using $\mathcal{A}_{\text{select}}$:
\STATE $S \gets \mathcal{A}_{\text{select}}(\{\ell_i\}_{i=1}^N, k)$

\vspace{0.5em}
\STATE \textbf{Simulate ground truth for selected coreset:}
\STATE $D_c \gets \emptyset$
\FOR{each $i \in S$}
    \STATE $u_i^\dagger \gets \mathcal{G}^\dagger(a_i)$ \COMMENT{Run numerical simulation}
    \STATE $D_c \gets D_c \cup \{(a_i, u_i^\dagger)\}$
\ENDFOR

\vspace{0.5em}
\STATE \textbf{Train} $\mathcal{G}_\theta$ on $D_c$ using supervised loss:
\FOR{$t = 1$ to $T$}
    \FOR{each $(a_i, u_i^\dagger) \in D_c$}
        \STATE $\theta \gets \theta - \alpha \nabla_\theta \mathcal{L}_{\text{data}}(a_i, u_i^\dagger; \theta)$
    \ENDFOR
\ENDFOR
\end{algorithmic}
\end{algorithm}

To address both issues of training time and data labeling costs for Neural Operator learning, we introduce PICore, an unsupervised coreset selection method that leverages a physics-informed loss to bypass the need for labeled training data during coreset selection.

Instead of using the ground truth PDE solution and supervised losses, the physics-informed loss evaluates the degree to which operator approximation \(\mathcal{G}_\theta(a)\) satisfies the governing PDEs defined in either the stationary form or the dynamic form. The physics-informed loss penalizes violations of the PDE (PDE residual) in the interior of the domain, as well as deviations from the given boundary and initial conditions. For neural operators, the physics-informed loss is defined as
\begin{equation}
\begin{aligned}
    \mathcal{L}_{PI}(a; \theta) = \bigg\|\mathcal{F}(\mathcal{G}_{\theta}(a), a)\bigg\|^2_{L^2(\Omega)} + \lambda \bigg\|\mathcal{G}_\theta(a)\big|_{\partial \Omega} - h\bigg\|^2_{L^2(\partial\Omega)}
\end{aligned}
\end{equation}
for stationary PDEs and 
\begin{equation}
\begin{aligned}
    \mathcal{L}_{PI}(a; \theta) = \bigg\|\mathcal{F}(\mathcal{G}_{\theta}(a), a)\bigg\|^2_{L^2(\Omega \times \mathcal{T})} + \lambda \bigg\|\mathcal{G}_\theta(a)\big|_{\partial \Omega \times \mathcal{T}} - h\bigg\|^2_{L^2(\partial\Omega \times \mathcal{T})} + \mu \bigg\|\mathcal{G}_\theta(a)\big|_{t=0} - a\bigg\|^2_{L^2(\Omega)}
\end{aligned}
\end{equation}
for dynamic PDEs. 

Given solely an unlabeled dataset \(D = \{a_i\}_{i=1}^N\) that can be cheaply generated (usually by sampling from a prior distribution or sensor readings), PICore selects a coreset of \(D\) by solving 
\begin{align}
    S = \argmin_{S' \subset D, |S'|=\beta N} \mathbb{E}_{a_i \sim S'} \left[\mathcal{L}_{PI}\left(a_i; \theta^{S'}\right)\right]
\end{align}
using any existing coreset selection algorithm where \(\theta^{S'}\) is the operator trained on \(S'\). After selecting the coreset \(S\), we simulate the true solutions \(u_i^{\dagger} = \mathcal{G}(a_i)\) for each \(a_i \in S\) using a traditional numerical solver, which forms the labeled subset \(D_{c} = \{(a_i, u_i^{\dagger})\}_{a_i \in S}\). Finally, we train the neural operator \(\mathcal{G}_\theta\) on \(D_c\) for \(T\) epochs with the standard supervised data loss
\begin{align}
    \mathcal{L}_{\text{data}}(a_i, u_i^\dagger) = \|\mathcal{G}_\theta(a_i) - u_i^{\dagger}\|_{L^2(\Omega \times \mathcal{T})}^2
\end{align}
Before coreset selection, we warm-start the neural operator with the physics-informed loss over the full dataset for a small number of epochs \(T_w << T\). Warm starting is common in prior coreset selection methods \citep{killamsetty2021grad} and is necessary as most coreset selection algorithms require gradient information, which is unusable with a randomly initialized model. We provide the full algorithm in Algorithm~\ref{alg:picore}.

\paragraph{Computing PDE Residuals}
One challenge with using the physics-informed loss in coreset selection is computing the PDE residual \(\mathcal{F}(\mathcal{G}_\theta(a), a)\). The residual requires computing derivatives of the neural operator with respect to the dimensional parameters, such as \(\frac{\partial^2 \mathcal{G}_\theta}{\partial x \partial t}\). \citet{li2024physics} uses a function-wise differentiation method via Fourier differentiation to compute these values exactly, but this does not extend to a general class of neural operators. We also tried auto-differentiation methods, but these were highly computationally expensive, increasing the coreset selection time. Thus, we settled on simply using finite difference methods, which are efficient with linear time complexity in the input resolution.

\section{Experimental Details}
\label{sec: experimental details}
We conduct experiments on four representative PDE benchmarks spanning both stationary and time-dependent dynamics widely used in the neural operator literature:
\begin{itemize}
    \item \textbf{1D Advection Equation} (time-dependent): A linear hyperbolic PDE representing pure transport dynamics, used to test propagation accuracy.
    \item \textbf{1D Burgers’ Equation} (time-dependent): A nonlinear convection-diffusion PDE with periodic boundary conditions, modeling shock formation and dissipation.
    \item \textbf{2D Darcy Flow} (stationary): A second-order elliptic PDE used to model pressure fields in porous media given heterogeneous permeability.
    \item \textbf{2D Navier-Stokes Incompressible Equation} (time-dependent): A nonlinear incompressible flow equation solved on a periodic domain.
\end{itemize}
\rebuttal{Each dataset has 1000 generated trajectories, with 900 that can be used for training (varying based on the coreset selection percentage) and 100 for testing, which is comparable to existing neural operator literature \citep{li2021fourierneuraloperatorparametric, li2024physics}. We generate 20 timesteps forward for Advection and Burgers, and only 10 timesteps for the Navier Stokes Incompressible dataset due to memory limits.} Additional information on the datasets can be found in Section~\ref{sec: PDE Datasets}.
We use the Fourier Neural Operator (FNO) \citep{li2020fourier} and U-Net Neural Operator \citep{rahman2023uno} as the base models for all experiments due to their implementation simplicity and performance. However, PICore can work out of the box with any neural operator. We also use 5 coreset selection algorithms in our experiments: CRAIG \citep{mirzasoleiman2020coresets}, GradMatch \citep{killamsetty2021grad}, AdaCore \citep{pooladzandi2022adaptive}, EL2N \citep{NEURIPS2021_ac56f8fe} and graNd \citep{NEURIPS2021_ac56f8fe}. \rebuttal{CRAIG, AdaCore, and GradMatch are submodular methods that try to match the gradient sum of the coreset to the gradient sum of the entire dataset. GraNd and EL2N are score based methods that use the gradient or the loss.} Additional information on these coreset selection algorithms can be found in Section~\ref{sec: coreset selection algorithms}. We use coreset selection percentages of 20\%, 30\%, 40\%, 60\%, and 80\%.

We report the results of each experiment with the normalized root mean square error loss (NRMSE):
\begin{align*}
    \frac{\| \mathcal{G}_\theta(a_i) - u_i^\dagger \|^2_{L^2(\Omega \times \mathcal{T})}}{\| u_i^\dagger \|^2_{L^2(\Omega \times \mathcal{T})}}
\end{align*}
used in \citet{PDEBench2022}. We use this as a normalized version of the data loss because the value of the \(u_i^\dagger\) at each spatiotemporal point is very small, resulting in small MSE values and potential gradient vanishing during training. We also use the uniform spatiotemporal discretization at an input resolution of 64 for \(\Omega\). Since FNO and UNO are resolution invariant, we also evaluate at higher resolutions for zero-shot super resolution in Section~\ref{subsec: zero shot super resolution}. For all experiments we use \(\lambda = 1\) and \(\mu = 1\), but this is relatively arbitrary, we did not conduct any hyperparameter tuning.

We use \(T_w = 25\) warmup epochs and reset the neural operator to its initialization to ensure fair comparisons between supervised and physics-informed coreset selection. Then, we train neural operators for \(T = 500\) epochs and report the average NRMSE over 5 seeds on a held-out test set at the input resolution. \rebuttal{We calculate the acceleration as the total time taken for supervised coreset selection / PICore (including data generation, warm starting, and training time) divided by the total time for the non-coreset baseline.}

\rebuttal{In addition to supervised coreset selection, we compare PICore to random subset selection, pure unsupervised training with the physics informed loss, and an active learning baseline based on uncertainty. Since most active learning baselines are for classification problems, we extend loss-as-uncertainty methods in \citet{liu2023understanding, 8579074} to neural operators. For the active learning baseline, we begin by randomly selecting 10\% of the available data as an initial training set and generating the corresponding ground-truth PDE solutions. We then train 10 independent copies of the neural operator on this subset for $T_w$ epochs, which is the same training time for the other methods. After training, we construct the final coreset from the remaining unlabeled data in a single step by selecting the points exhibiting the highest variance across the model predictions.}

\section{Results}
\subsection{Main Results}
We report the core findings for PICore and supervised coreset selection across the four representative PDE datasets in Tables \ref{tab:advection_64_combined}, \ref{tab:burgers_64_combined}, \ref{tab:darcy_64_combined}, and \ref{tab:navierstokesincompressible_64_combined}. In addition to the average test NRMSE over the best coreset selection algorithm for each method, we show the decrease in full training time (including data annotation costs through simulation) relative to the non-coreset selection baseline. Our results demonstrate that PICore consistently achieves competitive test performance compared to supervised coreset selection while providing substantial computational  gains, primarily by reducing expensive data annotation (simulation) costs during the coreset selection phase. 

\paragraph{PICore significantly improves training  through reduced simulation costs.}
Across four representative PDE datasets—Advection, Burgers, Darcy, and Navier-Stokes Incompressible—PICore consistently reduces the total training time by cutting down expensive simulation-based annotation. These  gains become especially significant as the complexity of the PDE increases: Across the four datasets, PICore achieves average training time reductions of 0.9\%, 9.8\%, 30.1\%, and 78.0\% compared to supervised coreset selection, calculated by averaging the relative acceleration improvements at each selection percentage (20\%, 30\%, 40\%, 60\%, and 80\%). For example, at a 20\% coreset size, PICore achieves a 5.01$\times$ speedup on Darcy Flow (vs. 2.24$\times$ for supervised methods) and a 5.00$\times$ speedup on Navier-Stokes (vs. 1.14$\times$) using UNO. 

\rebuttal{As shown in Tables~\ref{tab:picore_speedups_advection}, \ref{tab:picore_speedups_burgers}, \ref{tab:picore_speedups_darcy}, and \ref{tab:picore_speedups_navierstokes}, the relative contributions of training and data generation speedups vary by dataset difficulty. For simpler datasets such as Advection and Burgers,  gains are driven primarily by reductions in training time. For example, Advection achieves a 79.7\% improvement in training time but only a 1.47\% improvement in data generation time at the 20\% coreset level. In contrast, for more challenging datasets, the impact of training time reductions diminishes, while reductions in data generation time play a more significant role in overall  gains.} These results show that PICore scales well to high-dimensional scientific problems where data annotation costs dominate training.

\paragraph{PICore matches supervised coreset methods in test accuracy at reduced data budgets.}
Despite the substantial  improvements, PICore remains competitive with strong supervised baselines in terms of test NRMSE. For instance, at a 20\% coreset size, PICore achieves NRMSE values of $3.46\times 10^{-2}$ for Advection using FNO and $2.84\times 10^{-2}$ for Burgers using UNO, very close to the supervised method values of $3.42\times 10^{-2}$ and $2.93\times 10^{-2}$, respectively. This holds across different coreset sizes, with minor variations, showing that PICore can be as effective as supervised coreset selection, even at low coreset selection percentages. \rebuttal{In fact, many dataset, model, and selection percentages combinations show that PICore improves upon supervised coreset selection, but there is largely no significant change in accuracy between the two methods.} However, we observe that not all coreset selection algorithms perform equally as well. Due to the convexity assumptions and Hessian approximations with CRAIG \citep{mirzasoleiman2020coresets} and AdaCore \citep{pooladzandi2022adaptive}, they have higher NRMSE losses compared to the other algorithms. \rebuttal{Thus, the tables almost always report GradMatch, GraNd, or El2N as the coreset algorithm to use for PICore and supervised coreset selection.}

\paragraph{Coreset Selection methods outperform Random and Active Learning baselines on most datasets.}
\rebuttal{Random subset selection consistently underperforms relative to both PICore and subset selection, and this difference increases as we increase the complexity of the dataset and decrease the selection percentage.  For example, random selection has an nRMSE of \(2.74 \times 10^{-1}\) on the Navier Stokes Incompressible dataset at a 20\% coreset selection percentage, where as PICore has an nRMSE of \(1.12 \times 10^{-2}\) and Supervised Coreset Selection has an nRMSE of \(9.57 \times 10^{-2}\). This shows that using PDE specific information (either supervised loss or the physics informed loss) is necessary to achieve a more accurate solution with less data.
Interestingly, we see that active learning outperforms the PICore on the Advection dataset with UNO on medium coreset selection percentages (40-60\%) and on the Navier Stokes Incompressible dataset with FNO. However, it is much worse on all other dataset and model combinations by a considerable margin.}

\paragraph{Physics-informed training alone is insufficient to achieve high accuracy.}
\rebuttal{Training a neural operator solely with the physics-informed loss on the full unlabeled dataset yields significantly worse performance than all other baselines, despite requiring no simulation cost. For example, on the Navier--Stokes Incompressible dataset with FNO, the physics-informed-only baseline achieves an nRMSE of \(1.46 \times 10^{-0}\), compared to PICore’s \(1.25 \times 10^{-1}\) at 20\% coreset selection, a decrease of an order of magnitude. This large gap arises because the physics-informed loss is unstable as a standalone training objective and fails to capture fine-grained solution details without supervised guidance. In contrast, PICore leverages the physics-informed loss as a proxy for selecting informative samples, then trains on a small labeled subset with the supervised nRMSE loss, which mitachieving both higher accuracy and stability.}

\paragraph{There is a tradeoff between  and absolute test accuracy.}
While PICore offers strong performance and efficiency, one tradeoff is that the absolute test accuracy relative to training on \(100\%\) of the data is lower. For example, on the Advection dataset with FNO, the 100\% training baseline yields an NRMSE of $2.13 \times 10^{-2}$, while PICore at 20\% yields $3.77\times 10^{-2}$. However, this is an inherent tradeoff for all coreset selection algorithms, as the selected coreset simply contains less information for training. Additionally, this is not specific to PICore, as similar reductions in accuracy hold for supervised coreset selection. In practice, one may want to select a higher selection percentage, such as \(40\%\), which would yield higher accuracy (\(2.69 \times 10^{-2}\)) while still maintaining a competitive  gain (\(2.54\times\)).

\begin{table}[h]
\centering
\caption{Advection NRMSE at resolution 64}
\label{tab:advection_64_combined}
\resizebox{\textwidth}{!}{%
\begin{tabular}{llcccccc}
\toprule
Operator & Method & 20.0\% & 30.0\% & 40.0\% & 60.0\% & 80.0\% & 100.0\% \\
\midrule
\multirow{4}{*}{FNO}
 & Physics-Informed & $8.43 \pm 0.03 \times 10^{-1}$ & $8.57 \pm 0.03 \times 10^{-1}$ & $8.73 \pm 0.04 \times 10^{-1}$ & $8.98 \pm 0.02 \times 10^{-1}$ & $9.14 \pm 0.02 \times 10^{-1}$ & $9.26 \pm 0.03 \times 10^{-1}$ \\
 & & $(4.72\times)$ & $(3.06\times)$ & $(2.38\times)$ & $(1.59\times)$ & $(1.19\times)$ & $(0.96\times)$ \\
 & Random & $\textbf{3.39} \pm 0.07 \times 10^{-2}$ & $\textbf{2.89} \pm 0.03 \times 10^{-2}$ & $\underline{2.68} \pm 0.02 \times 10^{-2}$ & $2.47 \pm 0.03 \times 10^{-2}$ & $2.37 \pm 0.04 \times 10^{-2}$ & $2.22 \pm 0.05 \times 10^{-2}$ \\
 &  & $(5.10\times)$ & $(3.32\times)$ & $(2.56\times)$ & $(1.72\times)$ & $(1.28\times)$ & $(1.00\times)$ \\
 & Active Learning & $8.32 \pm 0.58 \times 10^{-2}$ & $6.29 \pm 0.40 \times 10^{-2}$ & $4.78 \pm 0.27 \times 10^{-2}$ & $3.51 \pm 0.16 \times 10^{-2}$ & $2.96 \pm 0.07 \times 10^{-2}$ & $2.22 \pm 0.05 \times 10^{-2}$ \\
 &  & $(5.04\times)$ & $(3.28\times)$ & $(2.52\times)$ & $(1.69\times)$ & $(1.26\times)$ & $(1.00\times)$ \\
 & Supervised (graNd) & $\underline{3.42} \pm 0.12 \times 10^{-2}$ & $\underline{2.96} \pm 0.09 \times 10^{-2}$ & $\textbf{2.64} \pm 0.03 \times 10^{-2}$ & $\underline{2.42} \pm 0.03 \times 10^{-2}$ & $\underline{2.25} \pm 0.02 \times 10^{-2}$ & $2.22 \pm 0.05 \times 10^{-2}$ \\
 &  & $(4.70\times)$ & $(3.15\times)$ & $(2.45\times)$ & $(1.66\times)$ & $(1.26\times)$ & $(1.00\times)$ \\
 & PICore (graNd) & $3.46 \pm 0.13 \times 10^{-2}$ & $3.04 \pm 0.15 \times 10^{-2}$ & $2.69 \pm 0.05 \times 10^{-2}$ & $\textbf{2.40} \pm 0.04 \times 10^{-2}$ & $\textbf{2.25} \pm 0.04 \times 10^{-2}$ & $2.22 \pm 0.05 \times 10^{-2}$ \\
 &  & $(5.06\times)$ & $(3.27\times)$ & $(2.54\times)$ & $(1.68\times)$ & $(1.26\times)$ & $(1.00\times)$ \\
\midrule
\multirow{4}{*}{UNO}
& Physics-Informed & $8.07 \pm 0.05 \times 10^{-1}$ & $8.19 \pm 0.06 \times 10^{-1}$ & $8.28 \pm 0.05 \times 10^{-1}$ & $8.46 \pm 0.18 \times 10^{-1}$ & $9.09 \pm 0.37 \times 10^{-1}$ & $9.25 \pm 0.30 \times 10^{-1}$ \\
 & & $(4.87\times)$ & $(3.21\times)$ & $(2.46\times)$ & $(1.65\times)$ & $(1.23\times)$ & $(0.98\times)$ \\
 & Random & $1.59 \pm 0.02 \times 10^{-1}$ & $1.50 \pm 0.01 \times 10^{-1}$ & $1.44 \pm 0.007 \times 10^{-1}$ & $1.42 \pm 0.12 \times 10^{-1}$ & $1.32 \pm 0.12 \times 10^{-1}$ & $7.27 \pm 0.28 \times 10^{-2}$ \\
 &  & $(5.08\times)$ & $(3.35\times)$ & $(2.55\times)$ & $(1.70\times)$ & $(1.28\times)$ & $(1.00\times)$ \\
 & Active Learning & $1.96 \pm 0.04 \times 10^{-1}$ & $1.59 \pm 0.05 \times 10^{-1}$ & $\textbf{9.20} \pm 0.73 \times 10^{-2}$ & $\textbf{7.49} \pm 0.47 \times 10^{-2}$ & $\textbf{6.83} \pm 0.05 \times 10^{-2}$ & $7.27 \pm 0.28 \times 10^{-2}$ \\
 &  & $(5.05\times)$ & $(3.32\times)$ & $(2.52\times)$ & $(1.68\times)$ & $(1.26\times)$ & $(1.00\times)$ \\
 & Supervised (gradmatch) & $\textbf{1.55} \pm 0.02 \times 10^{-1}$ & $\underline{1.48} \pm 0.01 \times 10^{-1}$ & $\underline{1.42} \pm 0.02 \times 10^{-1}$ & $\underline{1.17} \pm 0.14 \times 10^{-1}$ & $\underline{8.69} \pm 1.23 \times 10^{-2}$ & $7.27 \pm 0.28 \times 10^{-2}$ \\
 &  & $(4.84\times)$ & $(3.23\times)$ & $(2.47\times)$ & $(1.67\times)$ & $(1.25\times)$ & $(1.00\times)$ \\
 & PICore (gradmatch) & $\textbf{1.55} \pm 0.01 \times 10^{-1}$ & $\textbf{1.47} \pm 0.01 \times 10^{-1}$ & $1.43 \pm 0.008 \times 10^{-1}$ & $1.26 \pm 0.09 \times 10^{-1}$ & $9.06 \pm 1.07 \times 10^{-2}$ & $7.27 \pm 0.28 \times 10^{-2}$ \\
 &  & $(5.07\times)$ & $(3.34\times)$ & $(2.53\times)$ & $(1.69\times)$ & $(1.26\times)$ & $(1.00\times)$ \\
\bottomrule
\end{tabular}
}
\end{table}

\begin{table}[h]
\centering
\caption{Burgers NRMSE at resolution 64}
\label{tab:burgers_64_combined}
\resizebox{\textwidth}{!}{%
\begin{tabular}{llcccccc}
\toprule
Operator & Method & 20.0\% & 30.0\% & 40.0\% & 60.0\% & 80.0\% & 100.0\% \\
\midrule
\multirow{4}{*}{FNO}
& Physics-Informed & $5.26 \pm 0.12 \times 10^{-1}$ & $4.55 \pm 0.06 \times 10^{-1}$ & $4.36 \pm 0.07 \times 10^{-1}$ & $4.23 \pm 0.03 \times 10^{-1}$ & $4.19 \pm 0.03 \times 10^{-1}$ & $4.12 \pm 0.01 \times 10^{-1}$ \\
 & & $(5.36\times)$ & $(3.45\times)$ & $(2.70\times)$ & $(1.80\times)$ & $(1.34\times)$ & $(1.07\times)$ \\
 & Random & $1.85 \pm 0.09 \times 10^{-2}$ & $1.18 \pm 0.06 \times 10^{-2}$ & $8.23 \pm 0.21 \times 10^{-3}$ & $5.82 \pm 0.21 \times 10^{-3}$ & $4.75 \pm 0.13 \times 10^{-3}$ & $3.95 \pm 0.10 \times 10^{-3}$ \\
 &  & $(5.07\times)$ & $(3.31\times)$ & $(2.56\times)$ & $(1.72\times)$ & $(1.28\times)$ & $(1.00\times)$ \\
 & Active Learning & $8.76 \pm 2.52 \times 10^{-2}$ & $4.57 \pm 0.95 \times 10^{-2}$ & $3.30 \pm 0.57 \times 10^{-2}$ & $2.06 \pm 0.07 \times 10^{-2}$ & $1.37 \pm 0.18 \times 10^{-2}$ & $3.95 \pm 0.10 \times 10^{-3}$ \\
 & & $(5.03\times)$ & $(3.25\times)$ & $(2.52\times)$ & $(1.68\times)$ & $(1.26\times)$ &  $(1.00\times)$\\
 & Supervised (gradmatch) & $\textbf{1.71} \pm 0.16 \times 10^{-2}$ & $\underline{1.12} \pm 0.09 \times 10^{-2}$ & $\textbf{7.68} \pm 0.29 \times 10^{-3}$ & $\textbf{5.24} \pm 0.14 \times 10^{-3}$ & $\underline{4.13} \pm 0.08 \times 10^{-3}$ & $3.95 \pm 0.10 \times 10^{-3}$ \\
 &  & $(3.28\times)$ & $(2.52\times)$ & $(2.11\times)$ & $(1.55\times)$ & $(1.22\times)$ & $(1.00\times)$ \\
 & PICore (el2n) & $\underline{1.81} \pm 0.08 \times 10^{-2}$ & $\textbf{1.12} \pm 0.07 \times 10^{-2}$ & $\underline{8.07} \pm 0.33 \times 10^{-3}$ & $\underline{5.49} \pm 0.08 \times 10^{-3}$ & $\textbf{4.07} \pm 0.10 \times 10^{-3}$ & $3.95 \pm 0.10 \times 10^{-3}$ \\
 &  & $(5.05\times)$ & $(3.30\times)$ & $(2.53\times)$ & $(1.68\times)$ & $(1.26\times)$ & $(1.00\times)$ \\
\midrule
\multirow{4}{*}{UNO}
 & Physics-Informed & $4.82 \pm 0.07 \times 10^{-1}$ & $4.56 \pm 0.06 \times 10^{-1}$ & $4.48 \pm 0.06 \times 10^{-1}$ & $4.61 \pm 0.03 \times 10^{-1}$ & $4.63 \pm 0.02 \times 10^{-1}$ & $4.65 \pm 0.04 \times 10^{-1}$ \\
 & & $(5.28\times)$ & $(3.46\times)$ & $(2.66\times)$ & $(1.77\times)$ & $(1.33\times)$ & $(1.06\times)$ \\
 & Random & $\underline{2.92} \pm 0.05 \times 10^{-2}$ & $2.55 \pm 0.05 \times 10^{-2}$ & $2.25 \pm 0.05 \times 10^{-2}$ & $1.83 \pm 0.03 \times 10^{-2}$ & $1.58 \pm 0.01 \times 10^{-2}$ & $1.49 \pm 0.04 \times 10^{-2}$ \\
 &  & $(4.99\times)$ & $(3.34\times)$ & $(2.55\times)$ & $(1.70\times)$ & $(1.27\times)$ & $(1.00\times)$ \\
 & Active Learning & $5.42 \pm 0.41 \times 10^{-2}$ & $4.12 \pm 0.11 \times 10^{-2}$ & $3.62 \pm 0.20 \times 10^{-2}$ & $3.03 \pm 0.23 \times 10^{-2}$ & $2.51 \pm 0.13 \times 10^{-2}$ & $1.49 \pm 0.04 \times 10^{-2}$ \\
 &  & $(5.01\times)$ & $(3.30\times)$ & $(2.51\times)$ & $(1.68\times)$ & $(1.26\times)$ & $(1.00\times)$ \\
 & Supervised (gradmatch) & $2.93 \pm 0.10 \times 10^{-2}$ & $\underline{2.42} \pm 0.06 \times 10^{-2}$ & $\underline{2.08} \pm 0.05 \times 10^{-2}$ & $\underline{1.73} \pm 0.03 \times 10^{-2}$ & $\textbf{1.54} \pm 0.02 \times 10^{-2}$ & $1.49 \pm 0.04 \times 10^{-2}$ \\
 &  & $(3.77\times)$ & $(2.77\times)$ & $(2.23\times)$ & $(1.59\times)$ & $(1.23\times)$ & $(1.00\times)$ \\
 & PICore (graNd) & $\textbf{2.84} \pm 0.05 \times 10^{-2}$ & $\textbf{2.36} \pm 0.05 \times 10^{-2}$ & $\textbf{2.06} \pm 0.04 \times 10^{-2}$ & $\textbf{1.72} \pm 0.05 \times 10^{-2}$ & $\underline{1.57} \pm 0.03 \times 10^{-2}$ & $1.49 \pm 0.04 \times 10^{-2}$ \\
 &  & $(5.05\times)$ & $(3.33\times)$ & $(2.52\times)$ & $(1.69\times)$ & $(1.26\times)$ & $(1.00\times)$ \\
\bottomrule
\end{tabular}
}
\end{table}

\begin{table}[h]
\centering
\caption{Darcy NRMSE at resolution 64}
\label{tab:darcy_64_combined}
\resizebox{\textwidth}{!}{%
\begin{tabular}{llcccccc}
\toprule
Operator & Method & 20.0\% & 30.0\% & 40.0\% & 60.0\% & 80.0\% & 100.0\% \\
\midrule
\multirow{4}{*}{FNO}
 & Physics-Informed & $1.46 \pm 0.003 \times 10^{0}$ & $1.47 \pm 0.003 \times 10^{0}$ & $1.47 \pm 0.002 \times 10^{0}$ & $1.47 \pm 0.001 \times 10^{0}$ & $1.47 \pm 0.002 \times 10^{0}$ & $1.48 \pm 0.001 \times 10^{0}$ \\
 & & $(7.43\times)$ & $(4.96\times)$ & $(3.75\times)$ & $(2.50\times)$ & $(1.88\times)$ & $(1.51\times)$ \\
 & Random & $1.34 \pm 0.03 \times 10^{-1}$ & $1.15 \pm 0.01 \times 10^{-1}$ & $9.99 \pm 0.11 \times 10^{-2}$ & $7.94 \pm 0.04 \times 10^{-2}$ & $7.07 \pm 0.16 \times 10^{-2}$ & $6.18 \pm 0.09 \times 10^{-2}$ \\
 &  & $(5.00\times)$ & $(3.36\times)$ & $(2.53\times)$ & $(1.69\times)$ & $(1.27\times)$ & $(1.00\times)$ \\
 & Active Learning & $2.01 \pm 0.16 \times 10^{-1}$ & $1.58 \pm 0.08 \times 10^{-1}$ & $1.25 \pm 0.06 \times 10^{-1}$ & $8.94 \pm 0.33 \times 10^{-2}$ & $7.19 \pm 0.25 \times 10^{-2}$ & $6.18 \pm 0.09 \times 10^{-2}$ \\
 &  & $(4.99\times)$ & $(3.31\times)$ & $(2.50\times)$ & $(1.66\times)$ & $(1.25\times)$ & $(1.00\times)$ \\
 & Supervised (el2n) & $\underline{1.26} \pm 0.01 \times 10^{-1}$ & $\textbf{1.07} \pm 0.007 \times 10^{-1}$ & $\textbf{9.43} \pm 0.09 \times 10^{-2}$ & $\underline{7.83} \pm 0.18 \times 10^{-2}$ & $\textbf{6.59} \pm 0.09 \times 10^{-2}$ & $6.18 \pm 0.09 \times 10^{-2}$ \\
 &  & $(1.98\times)$ & $(1.76\times)$ & $(1.59\times)$ & $(1.33\times)$ & $(1.14\times)$ & $(1.00\times)$ \\
 & PICore (el2n) & $\textbf{1.25} \pm 0.02 \times 10^{-1}$ & $\underline{1.12} \pm 0.02 \times 10^{-1}$ & $\underline{9.44} \pm 0.12 \times 10^{-2}$ & $\textbf{7.77} \pm 0.18 \times 10^{-2}$ & $\underline{6.84} \pm 0.18 \times 10^{-2}$ & $6.18 \pm 0.09 \times 10^{-2}$ \\
 &  & $(5.00\times)$ & $(3.32\times)$ & $(2.50\times)$ & $(1.67\times)$ & $(1.25\times)$ & $(1.00\times)$ \\
\midrule
\multirow{4}{*}{UNO}
 & Physics-Informed & $1.42 \pm 0.002 \times 10^{0}$ & $1.42 \pm 0.003 \times 10^{0}$ & $1.42 \pm 0.003 \times 10^{0}$ & $1.43 \pm 0.001 \times 10^{0}$ & $1.42 \pm 0.002 \times 10^{0}$ & $1.43 \pm 0.003 \times 10^{0}$ \\
 & & $(7.05\times)$ & $(4.71\times)$ & $(3.55\times)$ & $(2.37\times)$ & $(1.78\times)$ & $(1.46\times)$ \\
 & Random & $1.45 \pm 0.02 \times 10^{-1}$ & $1.22 \pm 0.03 \times 10^{-1}$ & $1.10 \pm 0.02 \times 10^{-1}$ & $9.23 \pm 0.22 \times 10^{-2}$ & $8.78 \pm 0.30 \times 10^{-2}$ & $7.57 \pm 0.13 \times 10^{-2}$ \\
 &  & $(5.03\times)$ & $(3.37\times)$ & $(2.53\times)$ & $(1.69\times)$ & $(1.27\times)$ & $(1.00\times)$ \\
 & Active Learning & $1.87 \pm 0.14 \times 10^{-1}$ & $1.54 \pm 0.08 \times 10^{-1}$ & $1.27 \pm 0.06 \times 10^{-1}$ & $1.02 \pm 0.03 \times 10^{-1}$ & $8.63 \pm 0.19 \times 10^{-2}$ & $7.57 \pm 0.13 \times 10^{-2}$ \\
 &  & $(5.04\times)$ & $(3.35\times)$ & $(2.52\times)$ & $(1.68\times)$ & $(1.26\times)$ & $(1.00\times)$ \\
 & Supervised (gradmatch) & $\textbf{1.28} \pm 0.03 \times 10^{-1}$ & $\underline{1.14} \pm 0.01 \times 10^{-1}$ & $\underline{9.84} \pm 0.16 \times 10^{-2}$ & $\underline{8.60} \pm 0.11 \times 10^{-2}$ & $\underline{7.70} \pm 0.10 \times 10^{-2}$ & $7.57 \pm 0.13 \times 10^{-2}$ \\
 &  & $(2.23\times)$ & $(1.93\times)$ & $(1.71\times)$ & $(1.38\times)$ & $(1.16\times)$ & $(1.00\times)$ \\
 & PICore (graNd) & $\textbf{1.28} \pm 0.03 \times 10^{-1}$ & $\textbf{1.12} \pm 0.01 \times 10^{-1}$ & $\textbf{9.67} \pm 0.16 \times 10^{-2}$ & $\textbf{8.42} \pm 0.14 \times 10^{-2}$ & $\textbf{7.61} \pm 0.11 \times 10^{-2}$ & $7.57 \pm 0.13 \times 10^{-2}$ \\
 &  & $(5.01\times)$ & $(3.33\times)$ & $(2.50\times)$ & $(1.67\times)$ & $(1.25\times)$ & $(1.00\times)$ \\
\bottomrule
\end{tabular}
}
\end{table}

\begin{table}[h]
\centering
\caption{Navier Stokes Incompressible NRMSE at resolution 64}
\label{tab:navierstokesincompressible_64_combined}
\resizebox{\textwidth}{!}{%
\begin{tabular}{llcccccc}
\toprule
Operator & Method & 20.0\% & 30.0\% & 40.0\% & 60.0\% & 80.0\% & 100.0\% \\
\midrule
\multirow{4}{*}{FNO}
 & Physics-Informed & $1.01 \pm 0.001 \times 10^{0}$ & $1.01 \pm 0.002 \times 10^{0}$ & $1.02 \pm 0.002 \times 10^{0}$ & $1.02 \pm 0.002 \times 10^{0}$ & $1.03 \pm 0.002 \times 10^{0}$ & $1.03 \pm 0.002 \times 10^{0}$ \\
 & & $(69.41\times)$ & $(46.30\times)$ & $(34.73\times)$ & $(23.29\times)$ & $(17.43\times)$ & $(13.32\times)$ \\
 & Random & $2.74 \pm 0.45 \times 10^{-1}$ & $5.59 \pm 0.80 \times 10^{-2}$ & $1.33 \pm 0.03 \times 10^{-2}$ & $9.06 \pm 0.24 \times 10^{-3}$ & $6.87 \pm 0.13 \times 10^{-3}$ & $5.66 \pm 0.11 \times 10^{-3}$ \\
 &  & $(5.00\times)$ & $(3.34\times)$ & $(2.50\times)$ & $(1.67\times)$ & $(1.25\times)$ & $(1.00\times)$ \\
 & Active Learning & $\textbf{9.32} \pm 6.96 \times 10^{-2}$ & $2.16 \pm 0.66 \times 10^{-2}$ & $1.27 \pm 0.03 \times 10^{-2}$ & $\textbf{7.86} \pm 0.14 \times 10^{-3}$ & $\textbf{6.24} \pm 0.21 \times 10^{-3}$ & $5.66 \pm 0.11 \times 10^{-3}$ \\
 &  & $(5.00\times)$ & $(3.33\times)$ & $(2.50\times)$ & $(1.67\times)$ & $(1.25\times)$ & $(1.00\times)$ \\
 & Supervised (el2n) & $\underline{9.57} \pm 3.87 \times 10^{-2}$ & $\textbf{1.75} \pm 0.07 \times 10^{-2}$ & $\textbf{1.18} \pm 0.04 \times 10^{-2}$ & $\underline{7.94} \pm 0.18 \times 10^{-3}$ & $\underline{6.28} \pm 0.16 \times 10^{-3}$ & $5.66 \pm 0.11 \times 10^{-3}$ \\
 &  & $(1.05\times)$ & $(1.05\times)$ & $(1.04\times)$ & $(1.03\times)$ & $(1.01\times)$ & $(1.00\times)$ \\
 & PICore (graNd) & $1.12 \pm 0.45 \times 10^{-1}$ & $\underline{1.81} \pm 0.12 \times 10^{-2}$ & $\underline{1.23} \pm 0.05 \times 10^{-2}$ & $8.00 \pm 0.23 \times 10^{-3}$ & $6.34 \pm 0.14 \times 10^{-3}$ & $5.66 \pm 0.11 \times 10^{-3}$ \\
 &  & $(5.00\times)$ & $(3.33\times)$ & $(2.50\times)$ & $(1.67\times)$ & $(1.25\times)$ & $(1.00\times)$ \\
\midrule
\multirow{4}{*}{UNO}
 & Physics-Informed & $1.01 \pm 0.001 \times 10^{0}$ & $1.03 \pm 0.004 \times 10^{0}$ & $1.03 \pm 0.001 \times 10^{0}$ & $1.03 \pm 0.001 \times 10^{0}$ & $1.04 \pm 0.002 \times 10^{0}$ & $1.04 \pm 0.003 \times 10^{0}$ \\
 & & $(29.42\times)$ & $(19.61\times)$ & $(14.74\times)$ & $(9.86\times)$ & $(7.38\times)$ & $(5.89\times)$ \\
 & Random & $2.72 \pm 0.04 \times 10^{-2}$ & $2.24 \pm 0.02 \times 10^{-2}$ & $1.95 \pm 0.01 \times 10^{-2}$ & $1.61 \pm 0.006 \times 10^{-2}$ & $\textbf{1.38} \pm 0.004 \times 10^{-2}$ & $1.24 \pm 0.004 \times 10^{-2}$ \\
 &  & $(5.02\times)$ & $(3.34\times)$ & $(2.51\times)$ & $(1.67\times)$ & $(1.25\times)$ & $(1.00\times)$ \\
 & Active Learning & $2.91 \pm 0.05 \times 10^{-2}$ & $2.36 \pm 0.03 \times 10^{-2}$ & $2.06 \pm 0.02 \times 10^{-2}$ & $1.61 \pm 0.01 \times 10^{-2}$ & $1.40 \pm 0.009 \times 10^{-2}$ & $1.24 \pm 0.004 \times 10^{-2}$  \\
 &  & $(5.00\times)$ & $(3.33\times)$ & $(2.50\times)$ & $(1.67\times)$ & $(1.25\times)$ & $(1.00\times)$ \\
 & Supervised (el2n) & $\underline{2.60} \pm 0.02 \times 10^{-2}$ & $\textbf{2.19} \pm 0.02 \times 10^{-2}$ & $\underline{1.93} \pm 0.01 \times 10^{-2}$ & $\textbf{1.57} \pm 0.009 \times 10^{-2}$ & $\textbf{1.38} \pm 0.010 \times 10^{-2}$ & $1.24 \pm 0.004 \times 10^{-2}$  \\
 &  & $(1.14\times)$ & $(1.12\times)$ & $(1.10\times)$ & $(1.07\times)$ & $(1.03\times)$ & $(1.00\times)$ \\
 & PICore (gradmatch) & $\textbf{2.59} \pm 0.03 \times 10^{-2}$ & $\underline{2.20} \pm 0.009 \times 10^{-2}$ & $\textbf{1.92} \pm 0.009 \times 10^{-2}$ & $\underline{1.60} \pm 0.005 \times 10^{-2}$ & $1.40 \pm 0.007 \times 10^{-2}$ & $1.24 \pm 0.004 \times 10^{-2}$  \\
 &  & $(5.00\times)$ & $(3.33\times)$ & $(2.50\times)$ & $(1.67\times)$ & $(1.25\times)$ & $(1.00\times)$ \\
\bottomrule
\end{tabular}
}
\end{table}
\begin{table}[h]
\centering
\begin{minipage}{0.48\textwidth}
\centering
\resizebox{\textwidth}{!}{%
\begin{tabular}{lcccccc}
\toprule
Coreset \% & \multicolumn{3}{c}{FNO} & \multicolumn{3}{c}{UNO} \\ \cmidrule(lr){2-4} \cmidrule(lr){5-7}
 & Train & Data & Warm-up & Train & Data & Warm-up \\ \midrule
20.0\% & +79.7\% & +1.4\% & -0.9\% & +80.4\% & +0.8\% & -4.5\% \\
30.0\% & +69.7\% & +1.2\% & -1.4\% & +70.7\% & +0.7\% & -4.5\% \\
40.0\% & +61.3\% & +1.1\% & -1.8\% & +61.7\% & +0.6\% & -4.5\% \\
60.0\% & +42.8\% & +0.7\% & -2.8\% & +43.2\% & +0.4\% & -4.5\% \\
80.0\% & +24.2\% & +0.4\% & -3.7\% & +24.2\% & +0.2\% & -4.5\% \\
\bottomrule
\end{tabular}%
}
\caption{Advection PICore component speedup.}
\label{tab:picore_speedups_advection}
\end{minipage}%
\hfill
\begin{minipage}{0.48\textwidth}
\centering
\resizebox{\textwidth}{!}{%
\begin{tabular}{lcccccc}
\toprule
Coreset \% & \multicolumn{3}{c}{FNO} & \multicolumn{3}{c}{UNO} \\ \cmidrule(lr){2-4} \cmidrule(lr){5-7}
 & Train & Data & Warm-up & Train & Data & Warm-up \\ \midrule
20.0\% & +70.3\% & +10.7\% & -0.8\% & +74.4\% & +6.6\% & -4.5\% \\
30.0\% & +61.5\% & +9.4\% & -1.3\% & +65.4\% & +5.8\% & -4.5\% \\
40.0\% & +54.0\% & +8.1\% & -1.6\% & +57.1\% & +5.0\% & -4.5\% \\
60.0\% & +37.7\% & +5.4\% & -2.4\% & +39.9\% & +3.3\% & -4.5\% \\
80.0\% & +21.4\% & +2.7\% & -3.3\% & +22.3\% & +1.7\% & -4.5\% \\
\bottomrule
\end{tabular}%
}
\caption{Burgers PICore component speedup.}
\label{tab:picore_speedups_burgers}
\end{minipage}
\end{table}

\begin{table}[h]
\centering
\begin{minipage}{0.48\textwidth}
\centering
\resizebox{\textwidth}{!}{%
\begin{tabular}{lcccccc}
\toprule
Coreset \% & \multicolumn{3}{c}{FNO} & \multicolumn{3}{c}{UNO} \\ \cmidrule(lr){2-4} \cmidrule(lr){5-7}
 & Train & Data & Warm-up & Train & Data & Warm-up \\ \midrule
20.0\% & +50.2\% & +30.4\% & -0.6\% & +55.9\% & +24.8\% & -3.3\% \\
30.0\% & +44.2\% & +26.6\% & -0.9\% & +49.2\% & +21.7\% & -3.3\% \\
40.0\% & +38.4\% & +22.8\% & -1.2\% & +42.7\% & +18.6\% & -3.3\% \\
60.0\% & +26.7\% & +15.2\% & -1.8\% & +29.7\% & +12.4\% & -3.3\% \\
80.0\% & +14.9\% & +7.6\% & -2.4\% & +16.5\% & +6.2\% & -3.3\% \\
\bottomrule
\end{tabular}%
}
\caption{Darcy PICore component speedup.}
\label{tab:picore_speedups_darcy}
\end{minipage}%
\hfill
\begin{minipage}{0.48\textwidth}
\centering
\resizebox{\textwidth}{!}{%
\begin{tabular}{lcccccc}
\toprule
Coreset \% & \multicolumn{3}{c}{FNO} & \multicolumn{3}{c}{UNO} \\ \cmidrule(lr){2-4} \cmidrule(lr){5-7}
 & Train & Data & Warm-up & Train & Data & Warm-up \\ \midrule
20.0\% & +5.2\% & +74.9\% & -0.1\% & +12.7\% & +67.5\% & -0.8\% \\
30.0\% & +4.5\% & +65.5\% & -0.1\% & +11.2\% & +59.0\% & -0.8\% \\
40.0\% & +3.9\% & +56.2\% & -0.1\% & +9.7\% & +50.6\% & -0.8\% \\
60.0\% & +2.7\% & +37.4\% & -0.2\% & +6.7\% & +33.7\% & -0.8\% \\
80.0\% & +1.5\% & +18.7\% & -0.2\% & +3.7\% & +16.9\% & -0.8\% \\
\bottomrule
\end{tabular}%
}
\caption{Navier Stokes PICore component speedup.}
\label{tab:picore_speedups_navierstokes}
\end{minipage}
\end{table}

\subsection{Further Results}
In addition to investigating the efficacy of PICore, we also aim to answer the following questions: (1) How does PICore compare to existing unsupervised dataset selection methods? (2) How different are subsets selected by PICore compared to those selected by supervised coreset selection?

\subsubsection{Unsupervised Coreset Selection}
We compare PICore to three unsupervised coreset selection methods: k-means clustering, cosine similarity, and Herding \citep{chen2012super}. For k-means clustering we use \(k = \beta N\) clusters, and choose the data points closest to those clusters. For cosine similarity, we evaluate the cosine similarity between all pairs of points, and perform greedy selection to choose the coreset. We report direct comparison of the test NRMSE for both methods in Figures ~\ref{fig:advection_64_unsupervised}, \ref{fig:burgers_64_unsupervised}, \ref{fig:darcy_64_unsupervised}, and \ref{fig:navierstokesincompressible_64_unsupervised} in Section~\ref{subsec: unsupervised selection comparison}. The results show that PICore consistently matches or outperforms the unsupervised baselines across all tested coreset sizes (20\% to 80\%) and neural operator architectures (FNO and UNO). For instance, on the Advection dataset at 20\% coreset size, PICore with the EL2N algorithm achieves a test NRMSE of \(3.29 \times 10^{-2}\), outperforming cosine similarity \(3.39 \times 10^{-2}\) and herding \(3.46 \times 10^{-2}\). Similar patterns are observed on the other datasets, indicating that PICore’s selection strategy generalizes well across both time-dependent and stationary PDEs compared to other unsupervised coreset selection strategies. We also note that these trends hold across neural operator architectures, with PICore outperforming unsupervised methods with both FNO and UNO architectures. While FNO does consistently outperform UNO across datasets (except Navier Stokes Incompressible), this is due to the architecture differences and not due to PICore, as shown by the increase in NRMSE for UNO on the non-coreset baseline.

These results highlight the advantage of incorporating PDE-specific information into the subset selection process. While clustering and similarity-based approaches may cover the input space evenly or preserve diversity, they do not necessarily target the data points where the model struggles. In contrast, PICore explicitly focuses on where the model's performance is likely to decrease by using the PDE's residual, resulting in improved predictive accuracy under minimal training data.

\subsubsection{Convergence of Coreset Selection vs Active Learning}
\label{subsubsec: active learning convergence}
\rebuttal{A key distinction between coreset selection and active learning lies in their approach to data selection, which in turn affects their convergence speed. This iterative nature can be suboptimal, as it can lead to selecting redundant data points \citep{li2024survey}. While some works have shown superior convergence of active learning methods \citep{haimovich2024convergence}, these are under specific optimizer settings and in easier image classification domains.}

\begin{figure}
    \centering
    \includegraphics[width=0.7\linewidth]{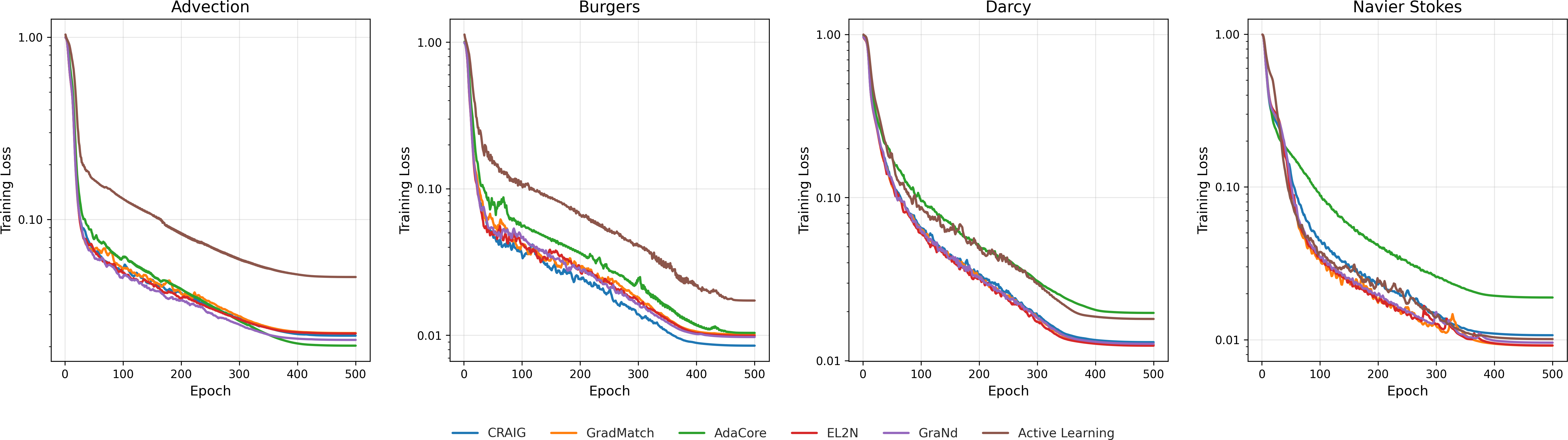}
    \caption{Training convergence of coreset selection v.s. active learning using FNO at a 20\% selection ratio.}
    \label{fig:fno_training_convergence}
\end{figure}

\begin{figure}
    \centering
    \includegraphics[width=0.7\linewidth]{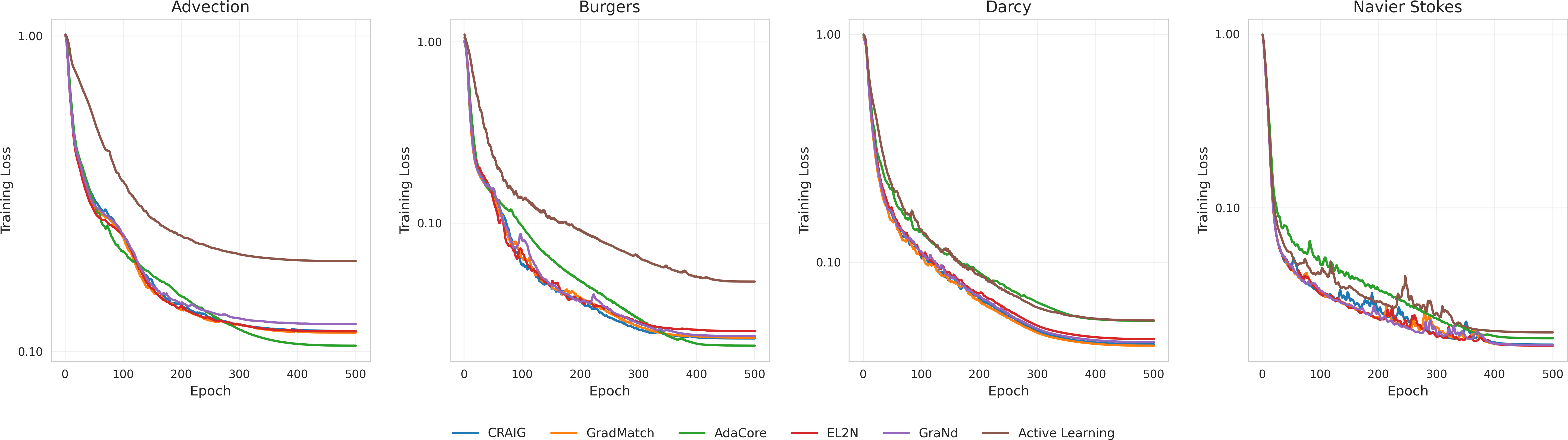}
    \caption{Training convergence of coreset selection v.s. active learning using UNO at a 20\% selection ratio.}
    \label{fig:uno_training_convergence}
\end{figure}

\rebuttal{Our empirical results largely validate this viewpoint, demonstrating that PICore's single-shot selection generally leads to better subset selection that converges faster than active learning baselines. Figures ~\ref{fig:fno_training_convergence} and ~\ref{fig:uno_training_convergence} show the training loss convergence of PICore's coreset selection methods compared to the active learning baseline. For both FNO and UNO, the active learning method converges much slower by 2-3\(\times\). The difference in loss convergence decreases for more complex datasets such as Navier Stokes, but this is due to learning a larger FNO-3D / UNO-3D model than due to the subset selection method itself.}
\section{Limitations and Future Work}
\rebuttal{One key limitation of PICore is the physics-informed loss requires that the underlying PDE is known and differentiable, which may not always be possible. However, practitioners can circumvent this by estimating the PDE analytically by using a trustworthy auxiliary model via domain knowledge to construct the PDE and using subdifferentials to estimate its derivatives, or numerically by employing data-driven surrogates or weak-form formulations that approximate the governing equations and their derivatives from observations. If such methods are not possible, alternative unsupervised criteria directly from observation, such as the distribution of the initial conditions, can be used with lower accuracy.} Another limitation of our work lies in its reliance on existing coreset selection algorithms. Methods such as CRAIG and AdaCore were originally developed under convexity assumptions and gradient and Hessian approximations and have shown strong performance in domains like image classification. However, applying such methods to complex PDE datasets, where the loss landscape is often highly non-convex, is less well understood. As a result, PICore’s performance at low selection ratios may be constrained by the suboptimal behavior of these algorithms. For example, using the Hutchinson Hessian approximation for AdaCore on only the last layer of the neural operator consistently results in poorer accuracies on the selected coresets than other coreset selection algorithms. Thus, practitioners are advised to use GradMatch, EL2N, or graNd as they operate under fewer model dependent assumptions and approximations. Future work could improve PICore by designing coreset selection algorithms tailored specifically for neural operator learning, by potentially leveraging inductive biases in operator networks. Additionally, while PICore can work with any neural operator architecture and input, we only use PICore on one input resolution with uniform geometries. Integrating multi-resolution data with arbitrary geometries could improve its generalization capability while maintaining efficiency.

\section{Conclusion}
In this work, we introduced PICore, a physics-informed unsupervised coreset selection framework designed to enhance the data efficiency of neural operator training. By leveraging the physics-informed loss to identify the most informative samples without requiring labeled data, PICore significantly reduces both the computational cost of numerical simulations and the time required for training. Our experiments across four PDE benchmarks demonstrate that PICore achieves competitive accuracy while reducing training costs by up to 78\% compared to supervised coreset selection methods. Although PICore inherits some limitations from existing selection methods, we believe its ability to reduce labeling costs and accelerate training makes it a promising tool for large-scale scientific machine learning.
\bibliography{main}

\begin{thebibliography}{55}
\providecommand{\natexlab}[1]{#1}
\providecommand{\url}[1]{\texttt{#1}}
\expandafter\ifx\csname urlstyle\endcsname\relax
  \providecommand{\doi}[1]{doi: #1}\else
  \providecommand{\doi}{doi: \begingroup \urlstyle{rm}\Url}\fi

\bibitem[Agarwal et~al.(2005)Agarwal, Har-Peled, Varadarajan, et~al.]{agarwal2005geometric}
Pankaj~K Agarwal, Sariel Har-Peled, Kasturi~R Varadarajan, et~al.
\newblock Geometric approximation via coresets.
\newblock \emph{Combinatorial and computational geometry}, 52\penalty0 (1):\penalty0 1--30, 2005.

\bibitem[Ash et~al.(2019)Ash, Zhang, Krishnamurthy, Langford, and Agarwal]{ash2019deep}
Jordan~T Ash, Chicheng Zhang, Akshay Krishnamurthy, John Langford, and Alekh Agarwal.
\newblock Deep batch active learning by diverse, uncertain gradient lower bounds.
\newblock \emph{arXiv preprint arXiv:1906.03671}, 2019.

\bibitem[Beluch et~al.(2018)Beluch, Genewein, Nurnberger, and Kohler]{8579074}
William~H. Beluch, Tim Genewein, Andreas Nurnberger, and Jan~M. Kohler.
\newblock The power of ensembles for active learning in image classification.
\newblock In \emph{2018 IEEE/CVF Conference on Computer Vision and Pattern Recognition}, pp.\  9368--9377, 2018.
\newblock \doi{10.1109/CVPR.2018.00976}.

\bibitem[Bonev et~al.(2023)Bonev, Kurth, Hundt, Pathak, Baust, Kashinath, and Anandkumar]{bonev2023spherical}
Boris Bonev, Thorsten Kurth, Christian Hundt, Jaideep Pathak, Maximilian Baust, Karthik Kashinath, and Anima Anandkumar.
\newblock Spherical fourier neural operators: Learning stable dynamics on the sphere.
\newblock In \emph{International conference on machine learning}, pp.\  2806--2823. PMLR, 2023.

\bibitem[Cao \& Tsang(2022)Cao and Tsang]{9234504}
Xiaofeng Cao and Ivor~W. Tsang.
\newblock Shattering distribution for active learning.
\newblock \emph{IEEE Transactions on Neural Networks and Learning Systems}, 33\penalty0 (1):\penalty0 215--228, 2022.
\newblock \doi{10.1109/TNNLS.2020.3027605}.

\bibitem[Chen et~al.(2024)Chen, Song, Ren, Subramanian, Morozov, and Mahoney]{chen2024data}
Wuyang Chen, Jialin Song, Pu~Ren, Shashank Subramanian, Dmitriy Morozov, and Michael~W Mahoney.
\newblock Data-efficient operator learning via unsupervised pretraining and in-context learning.
\newblock \emph{Advances in Neural Information Processing Systems}, 37:\penalty0 6213--6245, 2024.

\bibitem[Chen et~al.(2012)Chen, Welling, and Smola]{chen2012super}
Yutian Chen, Max Welling, and Alex Smola.
\newblock Super-samples from kernel herding.
\newblock \emph{arXiv preprint arXiv:1203.3472}, 2012.

\bibitem[Cyrus et~al.(1968)Cyrus, Fulton, Aeronautics, Administration, and Center]{cyrus1968accuracy}
N.J. Cyrus, R.E. Fulton, United States.~National Aeronautics, Space Administration, and Langley~Research Center.
\newblock \emph{Accuracy Study of Finite Difference Methods}.
\newblock NASA technical note. National Aeronautics and Space Administration, 1968.
\newblock URL \url{https://books.google.com/books?id=zMSFxfAasQMC}.

\bibitem[Eriksson \& Johnson(1995)Eriksson and Johnson]{eriksson_fem}
Kenneth Eriksson and Claes Johnson.
\newblock Adaptive finite element methods for parabolic problems iv: Nonlinear problems.
\newblock \emph{SIAM Journal on Numerical Analysis}, 32\penalty0 (6):\penalty0 1729--1749, 1995.
\newblock \doi{10.1137/0732078}.
\newblock URL \url{https://doi.org/10.1137/0732078}.

\bibitem[Gu et~al.(2021)Gu, Zhai, Deng, and Huang]{9178457}
Bin Gu, Zhou Zhai, Cheng Deng, and Heng Huang.
\newblock Efficient active learning by querying discriminative and representative samples and fully exploiting unlabeled data.
\newblock \emph{IEEE Transactions on Neural Networks and Learning Systems}, 32\penalty0 (9):\penalty0 4111--4122, 2021.
\newblock \doi{10.1109/TNNLS.2020.3016928}.

\bibitem[Guo et~al.(2022)Guo, Zhao, and Bai]{guo2022deepcore}
Chengcheng Guo, Bo~Zhao, and Yanbing Bai.
\newblock Deepcore: A comprehensive library for coreset selection in deep learning.
\newblock In \emph{International Conference on Database and Expert Systems Applications}, pp.\  181--195. Springer, 2022.

\bibitem[Gupta et~al.(2023)Gupta, Hasan, Prasad, and Gupta]{gupta2023data}
Animesh Gupta, Irtiza Hasan, Dilip~K Prasad, and Deepak~K Gupta.
\newblock Data-efficient training of cnns and transformers with coresets: A stability perspective.
\newblock \emph{arXiv preprint arXiv:2303.02095}, 2023.

\bibitem[Hacohen et~al.(2022)Hacohen, Dekel, and Weinshall]{hacohen2022activelearningbudgetopposite}
Guy Hacohen, Avihu Dekel, and Daphna Weinshall.
\newblock Active learning on a budget: Opposite strategies suit high and low budgets, 2022.
\newblock URL \url{https://arxiv.org/abs/2202.02794}.

\bibitem[Haimovich et~al.(2024)Haimovich, Karamshuk, Linder, Tax, and Vojnovic]{haimovich2024convergence}
Daniel Haimovich, Dima Karamshuk, Fridolin Linder, Niek Tax, and Milan Vojnovic.
\newblock On the convergence of loss and uncertainty-based active learning algorithms.
\newblock \emph{Advances in Neural Information Processing Systems}, 37:\penalty0 122770--122810, 2024.

\bibitem[Hemmasian \& Farimani(2024)Hemmasian and Farimani]{hemmasian2024pretraining}
AmirPouya Hemmasian and Amir~Barati Farimani.
\newblock Pretraining a neural operator in lower dimensions.
\newblock \emph{arXiv preprint arXiv:2407.17616}, 2024.

\bibitem[Houlsby et~al.(2011)Houlsby, Husz{\'a}r, Ghahramani, and Lengyel]{houlsby2011bayesian}
Neil Houlsby, Ferenc Husz{\'a}r, Zoubin Ghahramani, and M{\'a}t{\'e} Lengyel.
\newblock Bayesian active learning for classification and preference learning.
\newblock \emph{arXiv preprint arXiv:1112.5745}, 2011.

\bibitem[Johnson(1988)]{Johnson1988-em}
Claes Johnson.
\newblock \emph{Numerical solution of partial differential equations by the finite element method}.
\newblock Cambridge University Press, Cambridge, England, January 1988.

\bibitem[Killamsetty et~al.(2021{\natexlab{a}})Killamsetty, Durga, Ramakrishnan, De, and Iyer]{killamsetty2021grad}
Krishnateja Killamsetty, Sivasubramanian Durga, Ganesh Ramakrishnan, Abir De, and Rishabh Iyer.
\newblock Grad-match: Gradient matching based data subset selection for efficient deep model training.
\newblock In \emph{International Conference on Machine Learning}, pp.\  5464--5474. PMLR, 2021{\natexlab{a}}.

\bibitem[Killamsetty et~al.(2021{\natexlab{b}})Killamsetty, Sivasubramanian, Ramakrishnan, and Iyer]{killamsetty2021glister}
Krishnateja Killamsetty, Durga Sivasubramanian, Ganesh Ramakrishnan, and Rishabh Iyer.
\newblock Glister: Generalization based data subset selection for efficient and robust learning.
\newblock In \emph{Proceedings of the AAAI Conference on Artificial Intelligence}, volume~35, pp.\  8110--8118, 2021{\natexlab{b}}.

\bibitem[Killamsetty et~al.(2021{\natexlab{c}})Killamsetty, Zhao, Chen, and Iyer]{killamsetty2021retrieve}
Krishnateja Killamsetty, Xujiang Zhao, Feng Chen, and Rishabh Iyer.
\newblock Retrieve: Coreset selection for efficient and robust semi-supervised learning.
\newblock \emph{Advances in neural information processing systems}, 34:\penalty0 14488--14501, 2021{\natexlab{c}}.

\bibitem[Kim \& Shin(2022)Kim and Shin]{kim2022defense}
Yeachan Kim and Bonggun Shin.
\newblock In defense of core-set: A density-aware core-set selection for active learning.
\newblock In \emph{Proceedings of the 28th ACM SIGKDD conference on knowledge discovery and data mining}, pp.\  804--812, 2022.

\bibitem[Kim et~al.(2025)Kim, Kim, Ko, and Lee]{pmlr-v267-kim25m}
Yegon Kim, Hyunsu Kim, Gyeonghoon Ko, and Juho Lee.
\newblock Active learning with selective time-step acquisition for {PDE}s.
\newblock In Aarti Singh, Maryam Fazel, Daniel Hsu, Simon Lacoste-Julien, Felix Berkenkamp, Tegan Maharaj, Kiri Wagstaff, and Jerry Zhu (eds.), \emph{Proceedings of the 42nd International Conference on Machine Learning}, volume 267 of \emph{Proceedings of Machine Learning Research}, pp.\  30199--30223. PMLR, 13--19 Jul 2025.
\newblock URL \url{https://proceedings.mlr.press/v267/kim25m.html}.

\bibitem[Kovachki et~al.(2023)Kovachki, Li, Liu, Azizzadenesheli, Bhattacharya, Stuart, and Anandkumar]{kovachki2023neural}
Nikola Kovachki, Zongyi Li, Burigede Liu, Kamyar Azizzadenesheli, Kaushik Bhattacharya, Andrew Stuart, and Anima Anandkumar.
\newblock Neural operator: Learning maps between function spaces with applications to pdes.
\newblock \emph{Journal of Machine Learning Research}, 24\penalty0 (89):\penalty0 1--97, 2023.

\bibitem[Kovachki et~al.(2021)Kovachki, Li, Liu, Azizzadenesheli, Bhattacharya, Stuart, and Anandkumar]{DBLPjournalscorrabs-2108-08481}
Nikola~B. Kovachki, Zongyi Li, Burigede Liu, Kamyar Azizzadenesheli, Kaushik Bhattacharya, Andrew~M. Stuart, and Anima Anandkumar.
\newblock Neural operator: Learning maps between function spaces, 2021.
\newblock URL \url{https://arxiv.org/abs/2108.08481}.

\bibitem[LeVeque(2002)]{LeVeque_2002}
Randall~J. LeVeque.
\newblock \emph{Finite Volume Methods for Hyperbolic Problems}.
\newblock Cambridge Texts in Applied Mathematics. Cambridge University Press, 2002.

\bibitem[Li et~al.(2024{\natexlab{a}})Li, Wang, Chen, Jiang, Ding, and Okumura]{li2024survey}
Dongyuan Li, Zhen Wang, Yankai Chen, Renhe Jiang, Weiping Ding, and Manabu Okumura.
\newblock A survey on deep active learning: Recent advances and new frontiers.
\newblock \emph{IEEE Transactions on Neural Networks and Learning Systems}, 2024{\natexlab{a}}.

\bibitem[Li et~al.(2024{\natexlab{b}})Li, Yu, Xing, Kirby, Narayan, and Zhe]{li2024multi}
Shibo Li, Xin Yu, Wei Xing, Robert Kirby, Akil Narayan, and Shandian Zhe.
\newblock Multi-resolution active learning of fourier neural operators.
\newblock In \emph{International Conference on Artificial Intelligence and Statistics}, pp.\  2440--2448. PMLR, 2024{\natexlab{b}}.

\bibitem[Li et~al.(2020)Li, Kovachki, Azizzadenesheli, Liu, Bhattacharya, Stuart, and Anandkumar]{li2020fourier}
Zongyi Li, Nikola Kovachki, Kamyar Azizzadenesheli, Burigede Liu, Kaushik Bhattacharya, Andrew Stuart, and Anima Anandkumar.
\newblock Fourier neural operator for parametric partial differential equations.
\newblock \emph{arXiv preprint arXiv:2010.08895}, 2020.

\bibitem[Li et~al.(2021{\natexlab{a}})Li, Kovachki, Azizzadenesheli, Liu, Bhattacharya, Stuart, and Anandkumar]{li2021fourierneuraloperatorparametric}
Zongyi Li, Nikola Kovachki, Kamyar Azizzadenesheli, Burigede Liu, Kaushik Bhattacharya, Andrew Stuart, and Anima Anandkumar.
\newblock Fourier neural operator for parametric partial differential equations, 2021{\natexlab{a}}.
\newblock URL \url{https://arxiv.org/abs/2010.08895}.

\bibitem[Li et~al.(2021{\natexlab{b}})Li, Liu-Schiaffini, Kovachki, Liu, Azizzadenesheli, Bhattacharya, Stuart, and Anandkumar]{li2021learning}
Zongyi Li, Miguel Liu-Schiaffini, Nikola Kovachki, Burigede Liu, Kamyar Azizzadenesheli, Kaushik Bhattacharya, Andrew Stuart, and Anima Anandkumar.
\newblock Learning dissipative dynamics in chaotic systems.
\newblock \emph{arXiv preprint arXiv:2106.06898}, 2021{\natexlab{b}}.

\bibitem[Li et~al.(2022)Li, Liu-Schiaffini, Kovachki, Azizzadenesheli, Liu, Bhattacharya, Stuart, and Anandkumar]{li2022learning}
Zongyi Li, Miguel Liu-Schiaffini, Nikola~Borislavov Kovachki, Kamyar Azizzadenesheli, Burigede Liu, Kaushik Bhattacharya, Andrew Stuart, and Anima Anandkumar.
\newblock Learning chaotic dynamics in dissipative systems.
\newblock In Alice~H. Oh, Alekh Agarwal, Danielle Belgrave, and Kyunghyun Cho (eds.), \emph{Advances in Neural Information Processing Systems}, 2022.
\newblock URL \url{https://openreview.net/forum?id=1C36tFZn7sR}.

\bibitem[Li et~al.(2024{\natexlab{c}})Li, Zheng, Kovachki, Jin, Chen, Liu, Azizzadenesheli, and Anandkumar]{li2024physics}
Zongyi Li, Hongkai Zheng, Nikola Kovachki, David Jin, Haoxuan Chen, Burigede Liu, Kamyar Azizzadenesheli, and Anima Anandkumar.
\newblock Physics-informed neural operator for learning partial differential equations.
\newblock \emph{ACM/JMS Journal of Data Science}, 1\penalty0 (3):\penalty0 1--27, 2024{\natexlab{c}}.

\bibitem[Liu \& Li(2023)Liu and Li]{liu2023understanding}
Shang Liu and Xiaocheng Li.
\newblock Understanding uncertainty sampling.
\newblock \emph{arXiv preprint arXiv:2307.02719}, 2023.

\bibitem[Lu et~al.(2021)Lu, Jin, Pang, Zhang, and Karniadakis]{Lu_2021}
Lu~Lu, Pengzhan Jin, Guofei Pang, Zhongqiang Zhang, and George~Em Karniadakis.
\newblock Learning nonlinear operators via deeponet based on the universal approximation theorem of operators.
\newblock \emph{Nature Machine Intelligence}, 3\penalty0 (3):\penalty0 218–229, March 2021.
\newblock ISSN 2522-5839.
\newblock \doi{10.1038/s42256-021-00302-5}.
\newblock URL \url{http://dx.doi.org/10.1038/s42256-021-00302-5}.

\bibitem[Mirzasoleiman et~al.(2020)Mirzasoleiman, Bilmes, and Leskovec]{mirzasoleiman2020coresets}
Baharan Mirzasoleiman, Jeff Bilmes, and Jure Leskovec.
\newblock Coresets for data-efficient training of machine learning models.
\newblock In \emph{International Conference on Machine Learning}, pp.\  6950--6960. PMLR, 2020.

\bibitem[Musekamp et~al.(2024)Musekamp, Kalimuthu, Holzm{\"u}ller, Takamoto, and Niepert]{musekamp2024active}
Daniel Musekamp, Marimuthu Kalimuthu, David Holzm{\"u}ller, Makoto Takamoto, and Mathias Niepert.
\newblock Active learning for neural pde solvers.
\newblock \emph{arXiv preprint arXiv:2408.01536}, 2024.

\bibitem[Pathak et~al.(2022)Pathak, Subramanian, Harrington, Raja, Chattopadhyay, Mardani, Kurth, Hall, Li, Azizzadenesheli, Hassanzadeh, Kashinath, and Anandkumar]{pathak2022fourcastnet}
Jaideep Pathak, Shashank Subramanian, Peter Harrington, Sanjeev Raja, Ashesh Chattopadhyay, Morteza Mardani, Thorsten Kurth, David Hall, Zongyi Li, Kamyar Azizzadenesheli, Pedram Hassanzadeh, Karthik Kashinath, and Animashree Anandkumar.
\newblock Fourcastnet: A global data-driven high-resolution weather model using adaptive fourier neural operators.
\newblock \emph{arXiv preprint arXiv:2202.11214}, 2022.

\bibitem[Paul et~al.(2021)Paul, Ganguli, and Dziugaite]{NEURIPS2021_ac56f8fe}
Mansheej Paul, Surya Ganguli, and Gintare~Karolina Dziugaite.
\newblock Deep learning on a data diet: Finding important examples early in training.
\newblock In M.~Ranzato, A.~Beygelzimer, Y.~Dauphin, P.S. Liang, and J.~Wortman Vaughan (eds.), \emph{Advances in Neural Information Processing Systems}, volume~34, pp.\  20596--20607. Curran Associates, Inc., 2021.
\newblock URL \url{https://proceedings.neurips.cc/paper_files/paper/2021/file/ac56f8fe9eea3e4a365f29f0f1957c55-Paper.pdf}.

\bibitem[Pooladzandi et~al.(2022)Pooladzandi, Davini, and Mirzasoleiman]{pooladzandi2022adaptive}
Omead Pooladzandi, David Davini, and Baharan Mirzasoleiman.
\newblock Adaptive second order coresets for data-efficient machine learning.
\newblock In \emph{International Conference on Machine Learning}, pp.\  17848--17869. PMLR, 2022.

\bibitem[Rahman et~al.(2023)Rahman, Ross, and Azizzadenesheli]{rahman2023uno}
Md~Ashiqur Rahman, Zachary~E Ross, and Kamyar Azizzadenesheli.
\newblock U-{NO}: U-shaped neural operators.
\newblock \emph{Transactions on Machine Learning Research}, 2023.
\newblock ISSN 2835-8856.
\newblock URL \url{https://openreview.net/forum?id=j3oQF9coJd}.

\bibitem[Rahmati et~al.(2024)Rahmati, Fan, Zhou, Urban, Yoon, and Qian]{rahmati2024understanding}
Amir~Hossein Rahmati, Mingzhou Fan, Ruida Zhou, Nathan~M Urban, Byung-Jun Yoon, and Xiaoning Qian.
\newblock Understanding uncertainty-based active learning under model mismatch.
\newblock \emph{arXiv preprint arXiv:2408.13690}, 2024.

\bibitem[Raonic et~al.(2023)Raonic, Molinaro, De~Ryck, Rohner, Bartolucci, Alaifari, Mishra, and de~B{\'e}zenac]{raonic2023convolutional}
Bogdan Raonic, Roberto Molinaro, Tim De~Ryck, Tobias Rohner, Francesca Bartolucci, Rima Alaifari, Siddhartha Mishra, and Emmanuel de~B{\'e}zenac.
\newblock Convolutional neural operators for robust and accurate learning of pdes.
\newblock \emph{Advances in Neural Information Processing Systems}, 36:\penalty0 77187--77200, 2023.

\bibitem[Sener \& Savarese(2017)Sener and Savarese]{sener2017active}
Ozan Sener and Silvio Savarese.
\newblock Active learning for convolutional neural networks: A core-set approach.
\newblock \emph{arXiv preprint arXiv:1708.00489}, 2017.

\bibitem[Shim et~al.(2021)Shim, Kong, and Kang]{shim2021core}
Jae-hun Shim, Kyeongbo Kong, and Suk-Ju Kang.
\newblock Core-set sampling for efficient neural architecture search.
\newblock \emph{arXiv preprint arXiv:2107.06869}, 2021.

\bibitem[Sinha et~al.(2020)Sinha, Zhang, Goyal, Bengio, Larochelle, and Odena]{sinha2020small}
Samarth Sinha, Han Zhang, Anirudh Goyal, Yoshua Bengio, Hugo Larochelle, and Augustus Odena.
\newblock Small-gan: Speeding up gan training using core-sets.
\newblock In \emph{International Conference on Machine Learning}, pp.\  9005--9015. PMLR, 2020.

\bibitem[Takamoto et~al.(2022)Takamoto, Praditia, Leiteritz, MacKinlay, Alesiani, Pflüger, and Niepert]{PDEBench2022}
Makoto Takamoto, Timothy Praditia, Raphael Leiteritz, Dan MacKinlay, Francesco Alesiani, Dirk Pflüger, and Mathias Niepert.
\newblock {PDEBench: An Extensive Benchmark for Scientific Machine Learning}.
\newblock In \emph{36th Conference on Neural Information Processing Systems (NeurIPS 2022) Track on Datasets and Benchmarks}, 2022.
\newblock URL \url{https://arxiv.org/abs/2210.07182}.

\bibitem[Tang et~al.(2024)Tang, Kong, and Morris]{tang2024multi}
Hewei Tang, Qingkai Kong, and Joseph~P Morris.
\newblock Multi-fidelity fourier neural operator for fast modeling of large-scale geological carbon storage.
\newblock \emph{Journal of Hydrology}, 629:\penalty0 130641, 2024.

\bibitem[Tran et~al.(2021)Tran, Mathews, Xie, and Ong]{tran2021factorized}
Alasdair Tran, Alexander Mathews, Lexing Xie, and Cheng~Soon Ong.
\newblock Factorized fourier neural operators.
\newblock \emph{arXiv preprint arXiv:2111.13802}, 2021.

\bibitem[Wei et~al.(2015)Wei, Iyer, and Bilmes]{pmlr-v37-wei15}
Kai Wei, Rishabh Iyer, and Jeff Bilmes.
\newblock Submodularity in data subset selection and active learning.
\newblock In Francis Bach and David Blei (eds.), \emph{Proceedings of the 32nd International Conference on Machine Learning}, volume~37 of \emph{Proceedings of Machine Learning Research}, pp.\  1954--1963, Lille, France, 07--09 Jul 2015. PMLR.
\newblock URL \url{https://proceedings.mlr.press/v37/wei15.html}.

\bibitem[Welling(2009)]{10.1145/1553374.1553517}
Max Welling.
\newblock Herding dynamical weights to learn.
\newblock In \emph{Proceedings of the 26th Annual International Conference on Machine Learning}, ICML '09, pp.\  1121–1128, New York, NY, USA, 2009. Association for Computing Machinery.
\newblock ISBN 9781605585161.
\newblock \doi{10.1145/1553374.1553517}.
\newblock URL \url{https://doi.org/10.1145/1553374.1553517}.

\bibitem[Yang \& Loog(2022)Yang and Loog]{YANG2022108836}
Yazhou Yang and Marco Loog.
\newblock To actively initialize active learning.
\newblock \emph{Pattern Recognition}, 131:\penalty0 108836, 2022.
\newblock ISSN 0031-3203.
\newblock \doi{https://doi.org/10.1016/j.patcog.2022.108836}.
\newblock URL \url{https://www.sciencedirect.com/science/article/pii/S003132032200317X}.

\bibitem[Yao et~al.(2018)Yao, Xu, Roosta-Khorasani, and Mahoney]{yao2018inexactnonconvexnewtontypemethods}
Zhewei Yao, Peng Xu, Farbod Roosta-Khorasani, and Michael~W. Mahoney.
\newblock Inexact non-convex newton-type methods, 2018.
\newblock URL \url{https://arxiv.org/abs/1802.06925}.

\bibitem[Yoon et~al.(2021)Yoon, Madaan, Yang, and Hwang]{yoon2021online}
Jaehong Yoon, Divyam Madaan, Eunho Yang, and Sung~Ju Hwang.
\newblock Online coreset selection for rehearsal-based continual learning.
\newblock \emph{arXiv preprint arXiv:2106.01085}, 2021.

\bibitem[Zhang et~al.(2025)Zhang, Zhai, Ma, Shen, Li, Jiang, and Liu]{zhang2025staff}
Xiaoyu Zhang, Juan Zhai, Shiqing Ma, Chao Shen, Tianlin Li, Weipeng Jiang, and Yang Liu.
\newblock {STAFF}: Speculative coreset selection for task-specific fine-tuning.
\newblock In \emph{The Thirteenth International Conference on Learning Representations}, 2025.
\newblock URL \url{https://openreview.net/forum?id=FAfxvdv1Dy}.

\bibitem[Zhao et~al.(2021)Zhao, Dougherty, Yoon, Alexander, and Qian]{NEURIPS2021_50d2e70c}
Guang Zhao, Edward Dougherty, Byung-Jun Yoon, Francis Alexander, and Xiaoning Qian.
\newblock Efficient active learning for gaussian process classification by error reduction.
\newblock In M.~Ranzato, A.~Beygelzimer, Y.~Dauphin, P.S. Liang, and J.~Wortman Vaughan (eds.), \emph{Advances in Neural Information Processing Systems}, volume~34, pp.\  9734--9746. Curran Associates, Inc., 2021.
\newblock URL \url{https://proceedings.neurips.cc/paper_files/paper/2021/file/50d2e70cdf7dd05be85e1b8df3f8ced4-Paper.pdf}.

\end{thebibliography}
\bibliographystyle{tmlr}

\appendix
\section{PDE Datasets}
\label{sec: PDE Datasets}

For our experiments, we use several differential equation training sets to evaluate our algorithm. Each of these is used at an input grid resolution of 64. For the Advection, Burgers, and Darcy Flow equations, we generate datasets using code provided by \citet{PDEBench2022}. For the Navier-Stokes Incompressible equation dataset, we generate data from \citet{li2020fourier}.

\subsection{Advection}
We construct our dataset by numerically solving the linear advection equation on the periodic domain $(0,1)$:
\begin{equation}\label{eq:advection}
  \partial_t u(t,x) + \beta\,\partial_x u(t,x) = 0,
  \quad t \in (0,2],\; x \in (0,1),
\end{equation}
The initial condition is defined as a superposition of sinusoidal modes,
\begin{equation}\label{eq:advection_ic}
  u_0(x) \;=\;\sum_{i=1}^{N} A_i \,\sin\bigl(k_i x + \phi_i\bigr),
  \quad k_i = \frac{2\pi n_i}{L_x},
\end{equation}
where each $n_i$ is drawn uniformly from the range of integers from 1 to 8, $N$ is the number of waves, and the amplitudes $A_i\in[0,1]$ and phases $\phi_i\in(0,2\pi)$ are chosen at random. After assembly of $u_0(x)$, we apply with 10\% probability each a pointwise absolute‐value operation or multiplication by a smooth window function.

\subsection{Burger's Equation}

% In your LaTeX preamble, be sure to load:
%   \usepackage{amsmath,amssymb}

We are interested in the one-dimensional viscous Burgers equation on the unit interval with periodic boundary conditions:
\begin{equation}\label{eq:burgers_dataset}
  \partial_t u(t,x)
  + \partial_x\bigl(\tfrac12\,u^2(t,x)\bigr)
  = \frac{\nu}{\pi}\,\partial_{xx}u(t,x),
  \quad x\in(0,1),\; t\in(0,2],
\end{equation}
subject to the initial condition
\begin{equation}\label{eq:burgers_ic}
  u(0,x) = u_0(x), 
  \quad x\in(0,1).
\end{equation}
Here $\nu>0$ is a constant diffusion coefficient.  We use the nondimensional Reynolds number
\[
  R \;=\;\frac{\pi\,u_{L}}{\nu},
\]
where $u_L$ is a characteristic velocity scale.  In analogy with the Navier–Stokes equations, $R>1$ indicates a regime dominated by nonlinear steepening and potential shock formation, whereas $R<1$ corresponds to diffusion‐dominated smooth dynamics.

\subsection{Darcy Flow}

% In your LaTeX preamble, load amsmath and amssymb:
%   \usepackage{amsmath,amssymb}

We obtain the steady‐state solution of Darcy’s equation on the unit square by evolving a time‐dependent problem until convergence.  The target elliptic problem is
\begin{align}
  -\nabla\!\cdot\bigl(a(x)\,\nabla u(x)\bigr) &= f(x), 
  &x\in(0,1)^2,\label{eq:darcy_ss1}\\
  u(x)&=0, &x\in\partial(0,1)^2,\label{eq:darcy_ss2}
\end{align}
where $a(x)$ is the spatially varying coefficient and $f(x)\equiv\beta$ is a constant forcing that scales the solution amplitude.

Rather than solving \eqref{eq:darcy_ss1}, we integrate the parabolic problem
\begin{equation}\label{eq:darcy_transient}
  \partial_t u(x,t)
  - \nabla\!\cdot\bigl(a(x)\,\nabla u(x,t)\bigr)
  = \beta,
  \quad x\in(0,1)^2,\;t>0,
\end{equation}
with an appropriate random‐field initial condition and homogeneous Dirichlet boundary data. We use the strong form \(\nabla \cdot (a \nabla u) - f\) for the residual as in \citet{li2024physics}.

\subsection{Navier-Stokes Equation}

% In your LaTeX preamble, load amsmath and amssymb:
%   \usepackage{amsmath,amssymb}

We consider the vorticity formulation on the periodic domain $(0,1)^2$:
\[
\partial_t\omega + u\cdot\nabla\omega = \nu\,\Delta\omega + f,\quad
\nabla\!\cdot u = 0,\quad
\omega(x,0)\sim\mathcal{N}\bigl(0,\;7^{3/2}(-\Delta+49I)^{-2.5}\bigr),
\]
with forcing 
\[
f(x)=0.1\bigl[\sin2\pi(x_1+x_2)+\cos2\pi(x_1+x_2)\bigr].
\]

The solution is obtained on a $256\times256$ grid via a Fourier pseudospectral scheme: first, we solve $\Delta\psi=-\omega$ in Fourier space to recover the stream function $\psi$ and velocity $u$, then compute the nonlinear advection term $u\cdot\nabla\omega$ in physical space with a $2/3$‐dealiasing filter, and finally advance in time using Crank–Nicolson for diffusion coupled with an explicit update for the nonlinear term.

\section{Coreset Selection Algorithms}
\label{sec: coreset selection algorithms}
In this section, we provide an overview of the coreset selection algorithms used. All implementations are our own, but are based on \citet{guo2022deepcore}.

\subsection*{Adacore}
AdaCore augments CRAIG with second–order curvature so that difficult, high–influence samples are favoured even when first–order gradients look similar. In practice we \textbf{estimate only the diagonal} of the Hessian with \emph{10 Hutchinson probes} per mini-batch, then pre-condition the last-layer gradient $\nabla\ell_i$ by element-wise division. Similarities are computed on these pre-conditioned vectors and the same stochastic-greedy routine as CRAIG is applied.  The extra cost is the time to compute the approximation by deriving multiplications of the Hessian and arbitrary vectors via the Hessian-Free method \citep{yao2018inexactnonconvexnewtontypemethods}, the time of Hutchinson's method to find the diagonal, and the time to apply the diagonal to the gradients of the last layer.

\subsection*{EL2N}
Our EL2N (Error L2-Norm) coreset selection method follows from the premise that samples that are most worthwhile for the model have the highest losses. EL2N conducts a full training pass, where for each minibatch $x_i$, we calculate the loss without reduction for each individual sample, and calculate the norm for $x_i$'s loss vector. At the end of the epoch, we take the top $k$ minibatches by loss norm and return them with equal weight.

\subsection*{CRAIG}
CRAIG (\textbf{C}oresets for \textbf{A}ccelerating \textbf{I}ncremental \textbf{G}radient-descent) selects a weighted subset of size $k$ whose gradients cover (i.e.\ represent) all per-example gradients.  
Let $g_i=\nabla_{\!\theta}\ell_i(\theta)\in\mathbb R^{d}$ be the gradient for example~$i$.  
CRAIG finds a near optimal solution to the following problem. 
\[
A^* = \argmin_{A \subset V} |S|, \; \sum_{n \in V} \min_{m \in S} \max_{\theta} ||g_n - g_m||
\]
so every $g_i$ is “covered’’ by its most similar selected gradient. CRAIG selects the smallest subset S such that every example gradient is close (in $\mathcal{L}_2$) to at least one gradient in S. We approximate the coverage objective with the stochastic-greedy algorithm applied to the pairwise Euclidean similarity matrix of last-layer gradients.  
Greedy (or stochastic-greedy) selection gives a $(1-1/e)$-approximation in finite similarity evaluations.  
After $S$ is chosen, CRAIG sets integer weights  
\[
\gamma_j \;=\;\bigl|\{\,i : \argmax_{m\in S}s_{im}=j\}\bigr|,
\quad j\in S,
\]
so the weighted coreset gradient $\sum_{j\in S}\gamma_j g_j$ closely matches the full gradient $\sum_{i=1}^{n}g_i$ at each optimisation step.  
In practice the method is applied to last-layer gradients to reduce dimensionality without degrading the approximation quality.

\subsection*{GradMatch}

Let the last–layer per-example gradients be concatenated as
\(A=[g_1\, g_2\,\dots g_n]\in\mathbb{R}^{d\times n}\) and define the full-batch gradient
\(b=\tfrac1n\sum_{i=1}^{n}g_i\).
\textsc{GradMatch} casts coreset selection as the sparse approximation problem.
\[
  \min_{x\in\mathbb{R}^n}\;\|Ax-b\|_2^2
  \quad\text{s.t.}\quad
  \|x\|_0\le k,\;x\ge 0.
\]

OMP builds the weight vector \(x\) greedily.  
Starting with residual \(r=b\) and empty support \(S\):
(i) choose the column \(j^\star=\arg\max_{j\notin S}A_j^\top r\);
(ii) add \(j^\star\) to \(S\);
(iii) refit the coefficients by non-negative least squares
\(x_S=\arg\min_{x\ge0}\|A_Sx-b\|_2^2+\lambda\|x\|_2^2\);
(iv) update \(r=b-A_Sx_S\).
The loop terminates after \(k\) selections, giving a coreset
\(S=\operatorname{supp}(x)\) with weights \(\gamma_j=x_j\).

During training we replace the full loss by the weighted loss
\(\sum_{j\in S}\gamma_j\ell_j\big/\sum_{j\in S}\gamma_j\), ensuring the
mini-batch gradient of the coreset closely follows the full-batch
gradient throughout optimisation.

\subsection*{GraNd}
GraNd is similar to EL2N, but simply orders samples by the norm of their individual gradients and keeps the top $k$.  We piggy-back on the same per-sample gradient collection already needed for CRAIG/GradMatch, but \emph{stop after the first backward call}. We can rapidly sort these norms on the CPU, and use the selected indices for our coreset.

\section{Further Results}
\subsection{Unsupervised Selection Methods Comparison}
\label{subsec: unsupervised selection comparison}
\begin{figure}[htbp]
    \centering
    \includegraphics[width=\linewidth]{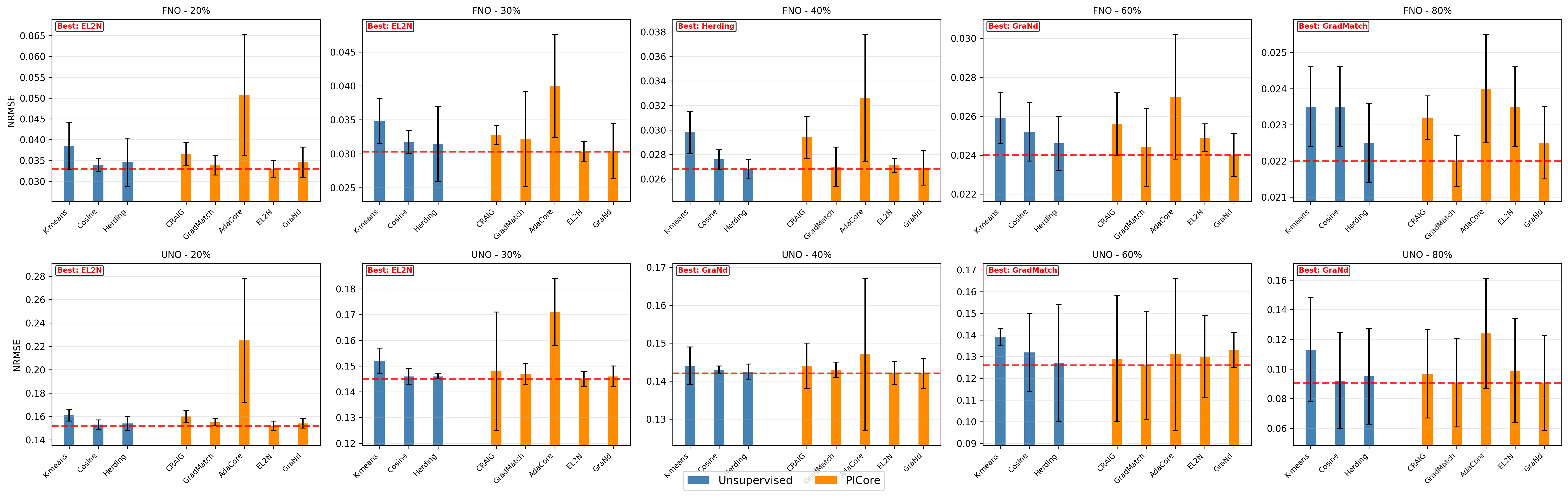}
    \caption{Test NRMSE on the Advection dataset at resolution 64 across varying coreset percentages (20\%–100\%) between unsupervised and PICore-based coreset selection methods using both FNO and UNO architectures.}
    \label{fig:advection_64_unsupervised}
\end{figure}

\begin{figure}[htbp]
    \centering
    \includegraphics[width=\linewidth]{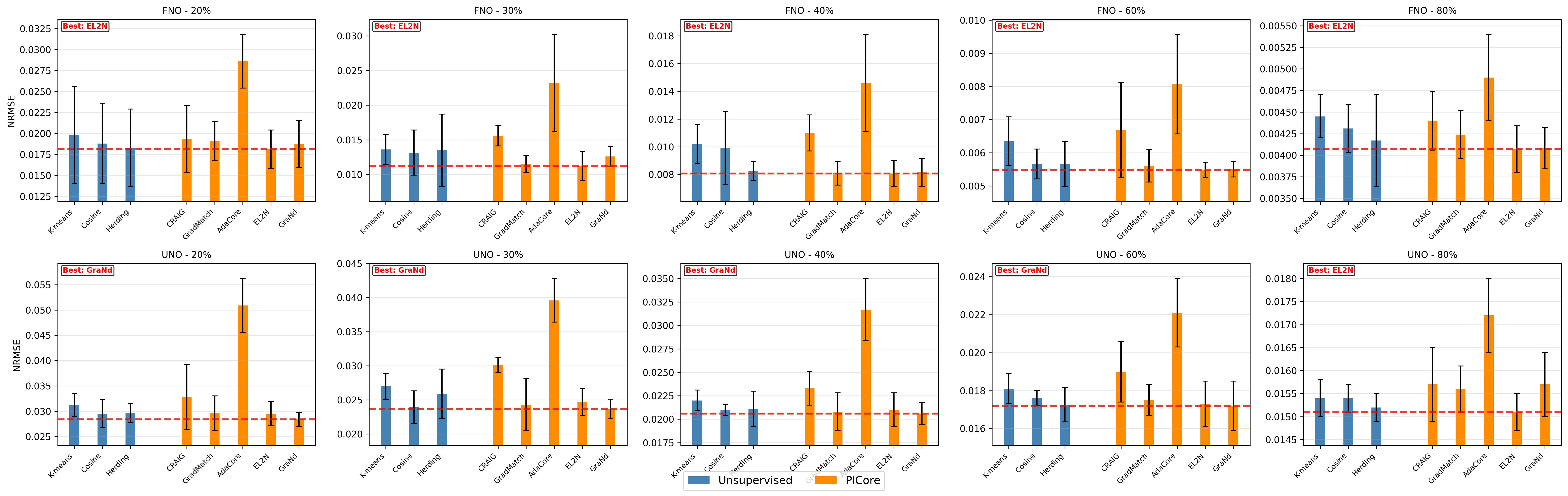}
    \caption{Test NRMSE on the Burgers dataset at resolution 64 across varying coreset percentages (20\%–100\%) between unsupervised and PICore-based coreset selection methods using both FNO and UNO architectures.}
    \label{fig:burgers_64_unsupervised}
\end{figure}

\begin{figure}[htbp]
    \centering
    \includegraphics[width=\linewidth]{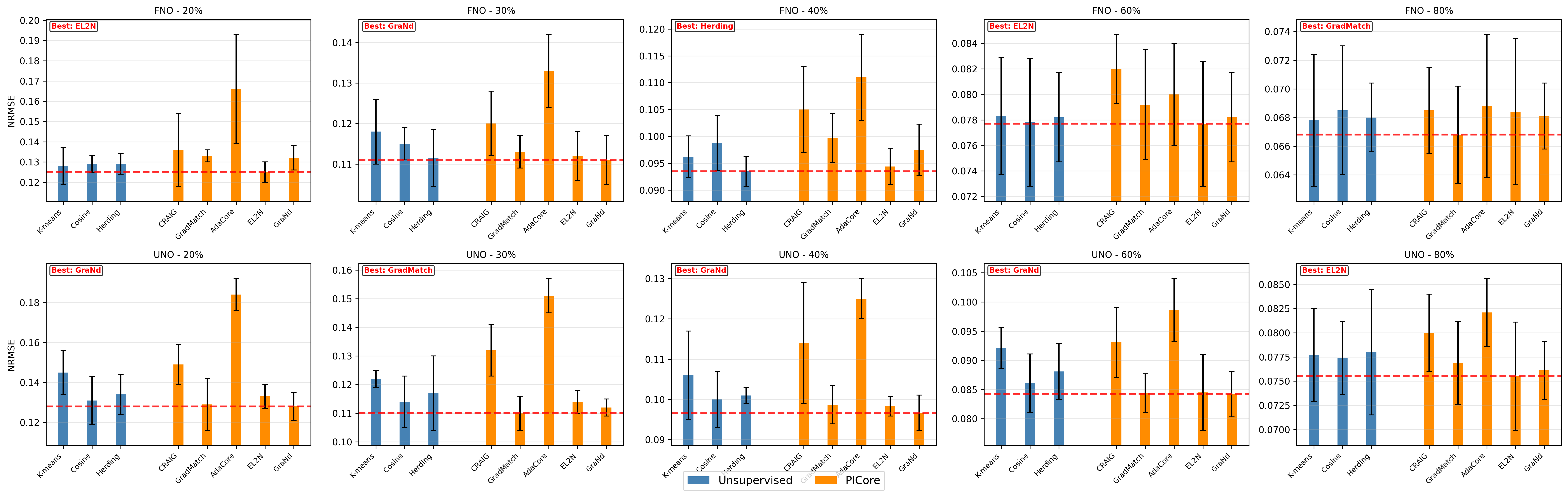}
    \caption{Test NRMSE on the Darcy dataset at resolution 64 across varying coreset percentages (20\%–100\%) between unsupervised and PICore-based coreset selection methods using both FNO and UNO architectures.}
    \label{fig:darcy_64_unsupervised}
\end{figure}

\begin{figure}[htbp]
    \centering
    \includegraphics[width=\linewidth]{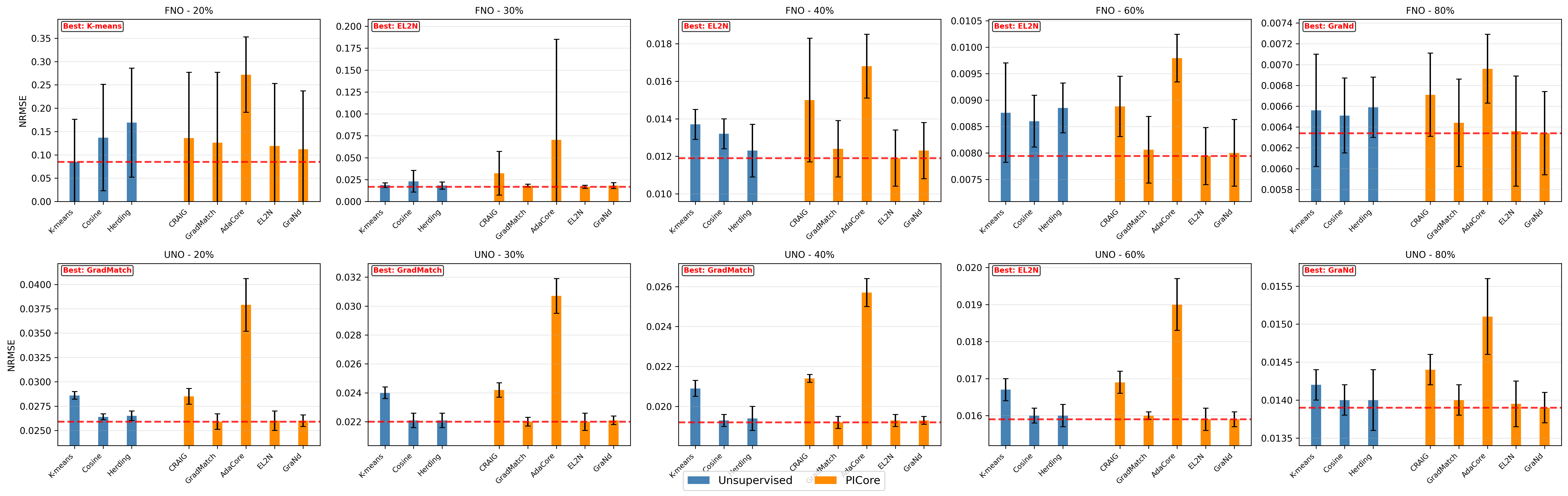}
    \caption{Test NRMSE on the Navier Stokes Incompressible dataset at resolution 64 across varying coreset percentages (20\%–100\%) between unsupervised and PICore-based coreset selection methods using both FNO and UNO architectures.}
    \label{fig:navierstokesincompressible_64_unsupervised}
\end{figure}

\clearpage

\subsection{Spatial Orientation of Supervised Coreset Selection and PICore}
\begin{figure}[h]
    \centering
    \begin{subfigure}[h]{0.35\textwidth}
        \centering
        \includegraphics[width=\textwidth]{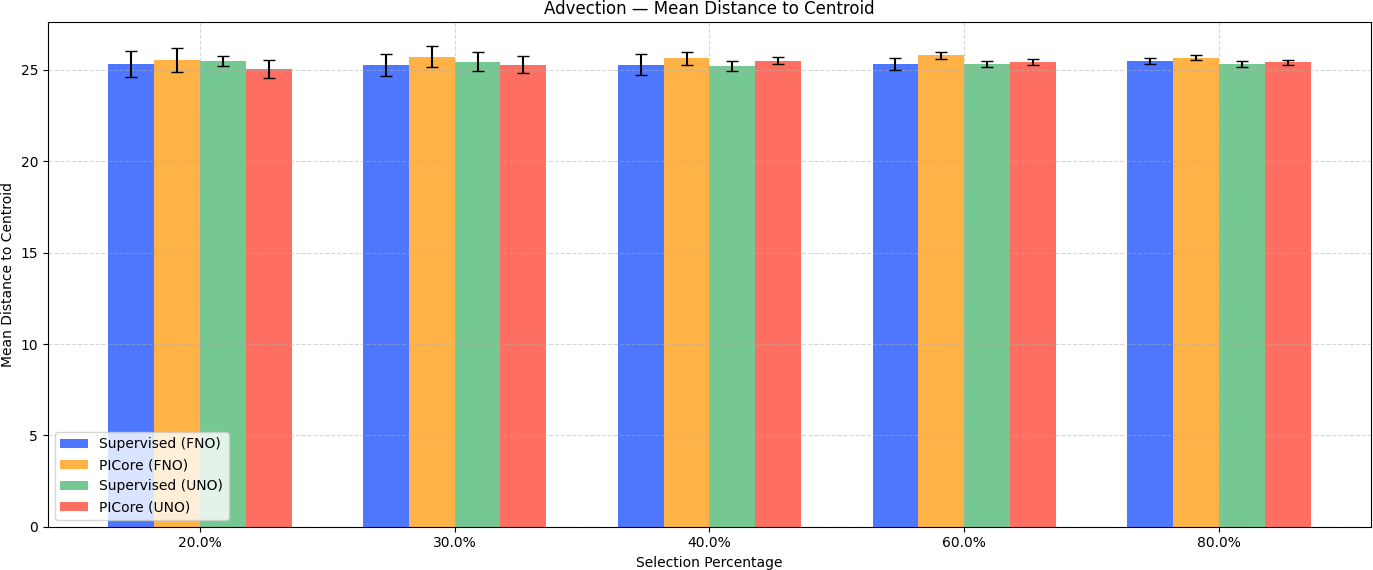}
        \caption{Advection}
    \end{subfigure}
    \begin{subfigure}[h]{0.35\textwidth}
        \centering
        \includegraphics[width=\textwidth]{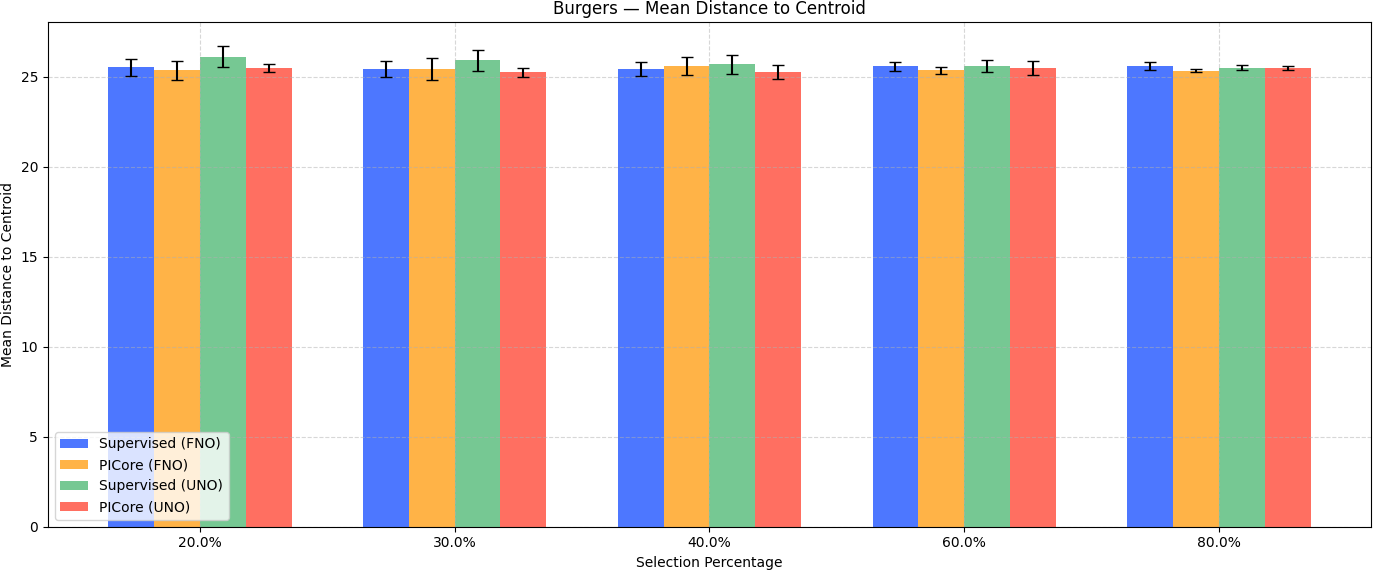}
        \caption{Burgers}
    \end{subfigure}
    \begin{subfigure}[h]{0.35\textwidth}
        \centering
        \includegraphics[width=\textwidth]{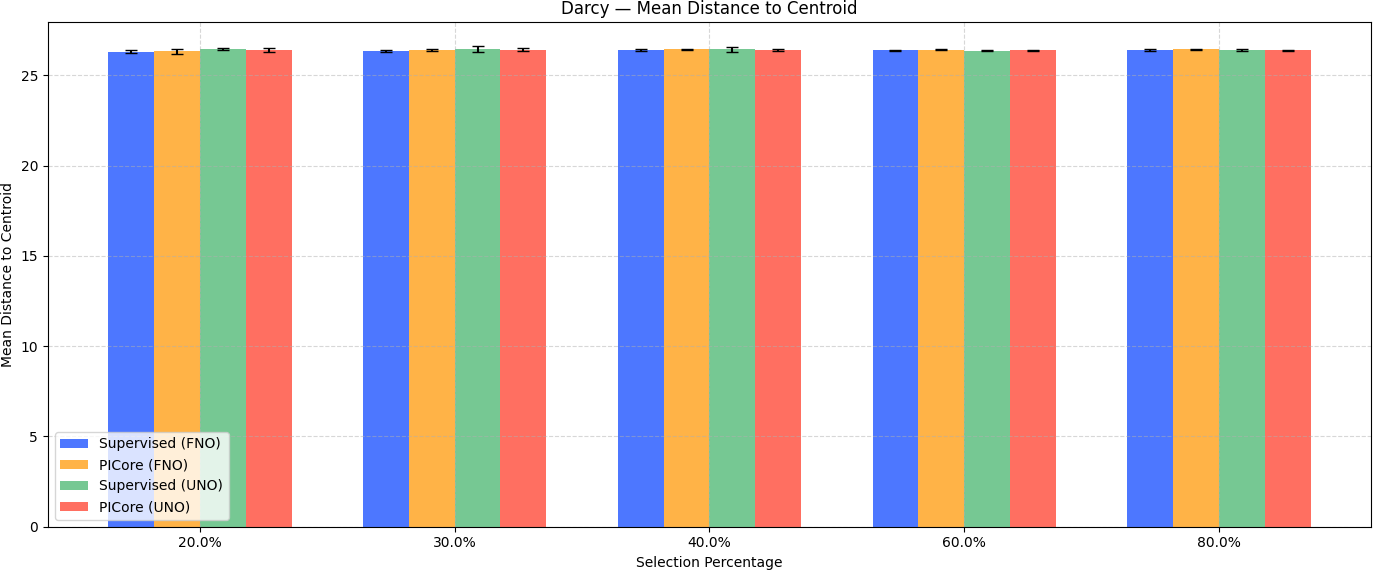}
        \caption{Darcy}
    \end{subfigure}
    \begin{subfigure}[h]{0.35\textwidth}
        \centering
        \includegraphics[width=\textwidth]{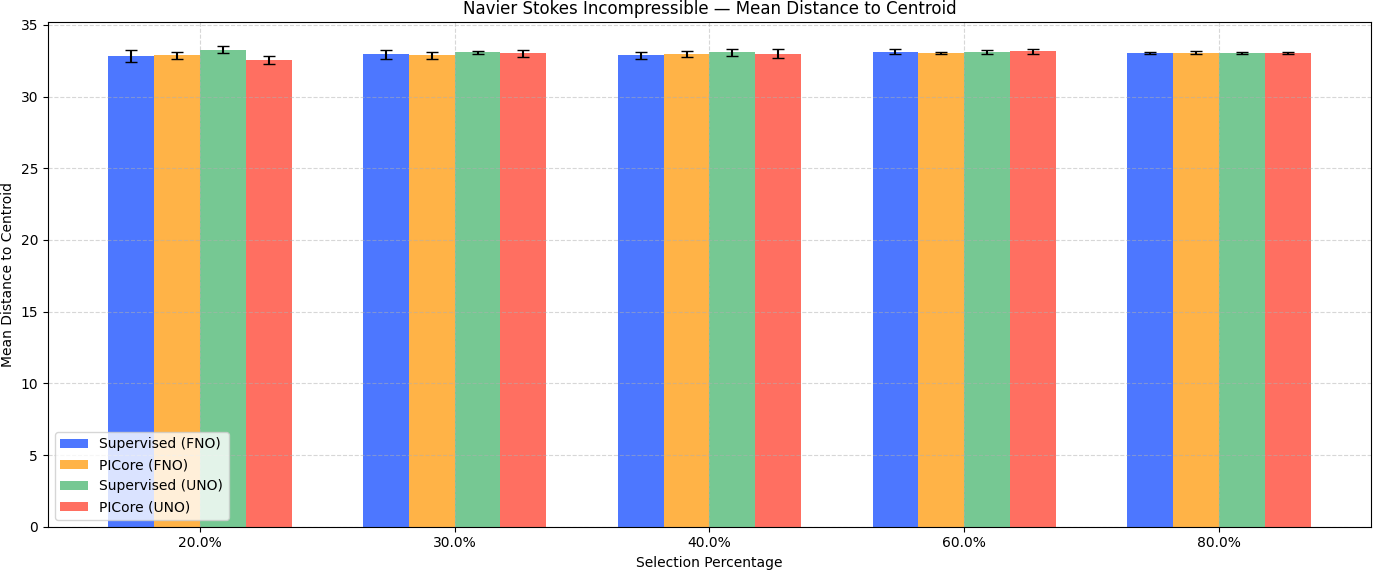}
        \caption{Navier Stokes}
    \end{subfigure}

    \caption{Average centroid distances across datasets for FNO and UNO.}
    \label{fig:fno_uno_centroids}
\end{figure}

To better understand the differences between supervised coreset selection and PICore, we analyze how well each method covers the input space by computing the average distance from coreset points to their centroid, which serves as a proxy for spread or diversity. \rebuttal{We compute this distance with respect to the \(\|\cdot\|_{L^2(\Omega)}\) norm, where the centroid is the average data point element-wise and the average distance is the average norm between the centroid and the selected data points in the coreset.} As shown in Figure~\ref{fig:fno_uno_centroids}, this distance is nearly identical across datasets and neural operators (FNO and UNO), with overlapping standard error bars with differences decreasing as the PDE complexity increases (Advection to Navier Stokes). This suggests that PICore selects coresets that are as well-distributed as those from supervised methods, despite not using labeled data. The comparable coverage indicates that differences in downstream performance likely arise from the type of points selected rather than their spatial distribution.

\subsection{Ablation Study}
\begin{table}[H]
\centering
\caption{Ablation on Warm Start at resolution 64}
\label{tab:burgers_64_ablation}
\resizebox{\textwidth}{!}{%
\begin{tabular}{llcccccc}
\toprule
Operator & $T_w$ & 20.0\% & 30.0\% & 40.0\% & 60.0\% & 80.0\% & 100.0\% \\
\midrule
\multirow{8}{*}{FNO}
 & 10 epochs & $1.64 \pm 0.14 \times 10^{-2}$ & $1.14 \pm 0.10 \times 10^{-2}$ & $8.07 \pm 0.48 \times 10^{-3}$ & $5.32 \pm 0.30 \times 10^{-3}$ & $4.17 \pm 0.17 \times 10^{-3}$ & $3.95 \pm 0.10 \times 10^{-3}$ \\
 &  & $(3.23\times)$ & $(2.48\times)$ & $(2.06\times)$ & $(1.52\times)$ & $(1.19\times)$ & $(1.00\times)$ \\
\cmidrule{2-8}
 & 25 epochs & $1.52 \pm 0.13 \times 10^{-2}$ & $1.07 \pm 0.06 \times 10^{-2}$ & $7.71 \pm 0.33 \times 10^{-3}$ & $5.33 \pm 0.14 \times 10^{-3}$ & $4.15 \pm 0.06 \times 10^{-3}$ & $3.95 \pm 0.10 \times 10^{-3}$ \\
 &  & $(3.18\times)$ & $(2.44\times)$ & $(2.02\times)$ & $(1.48\times)$ & $(1.17\times)$ & $(1.00\times)$ \\
\cmidrule{2-8}
 & 50 epochs & $1.69 \pm 0.07 \times 10^{-2}$ & $1.12 \pm 0.08 \times 10^{-2}$ & $7.96 \pm 0.38 \times 10^{-3}$ & $5.51 \pm 0.17 \times 10^{-3}$ & $4.16 \pm 0.11 \times 10^{-3}$ & $3.95 \pm 0.10 \times 10^{-3}$ \\
 &  & $(3.10\times)$ & $(2.36\times)$ & $(1.95\times)$ & $(1.43\times)$ & $(1.13\times)$ & $(1.00\times)$ \\
\cmidrule{2-8}
 & 100 epochs & $1.69 \pm 0.14 \times 10^{-2}$ & $1.06 \pm 0.04 \times 10^{-2}$ & $8.24 \pm 0.45 \times 10^{-3}$ & $5.40 \pm 0.16 \times 10^{-3}$ & $4.15 \pm 0.05 \times 10^{-3}$ & $3.95 \pm 0.10 \times 10^{-3}$ \\
 &  & $(2.94\times)$ & $(2.21\times)$ & $(1.83\times)$ & $(1.34\times)$ & $(1.05\times)$ & $(1.00\times)$ \\
\midrule
\multirow{8}{*}{UNO}
 & 10 epochs & $2.97 \pm 0.04 \times 10^{-2}$ & $2.44 \pm 0.06 \times 10^{-2}$ & $2.13 \pm 0.04 \times 10^{-2}$ & $1.77 \pm 0.04 \times 10^{-2}$ & $1.57 \pm 0.01 \times 10^{-2}$ & $1.49 \pm 0.04 \times 10^{-2}$ \\
 &  & $(3.79\times)$ & $(2.79\times)$ & $(2.25\times)$ & $(1.60\times)$ & $(1.24\times)$ & $(1.00\times)$ \\
\cmidrule{2-8}
 & 25 epochs & $2.99 \pm 0.04 \times 10^{-2}$ & $2.51 \pm 0.04 \times 10^{-2}$ & $2.09 \pm 0.02 \times 10^{-2}$ & $1.76 \pm 0.03 \times 10^{-2}$ & $1.57 \pm 0.04 \times 10^{-2}$ & $1.49 \pm 0.04 \times 10^{-2}$ \\
 &  & $(3.72\times)$ & $(2.73\times)$ & $(2.19\times)$ & $(1.56\times)$ & $(1.20\times)$ & $(1.00\times)$ \\
\cmidrule{2-8}
 & 50 epochs & $2.89 \pm 0.03 \times 10^{-2}$ & $2.42 \pm 0.03 \times 10^{-2}$ & $2.18 \pm 0.05 \times 10^{-2}$ & $1.73 \pm 0.03 \times 10^{-2}$ & $1.54 \pm 0.02 \times 10^{-2}$ & $1.49 \pm 0.04 \times 10^{-2}$ \\
 &  & $(3.60\times)$ & $(2.64\times)$ & $(2.11\times)$ & $(1.49\times)$ & $(1.15\times)$ & $(1.00\times)$ \\
\cmidrule{2-8}
 & 100 epochs & $3.00 \pm 0.08 \times 10^{-2}$ & $2.53 \pm 0.05 \times 10^{-2}$ & $2.05 \pm 0.04 \times 10^{-2}$ & $1.73 \pm 0.03 \times 10^{-2}$ & $1.55 \pm 0.02 \times 10^{-2}$ & $1.49 \pm 0.04 \times 10^{-2}$ \\
 &  & $(3.38\times)$ & $(2.46\times)$ & $(1.96\times)$ & $(1.38\times)$ & $(1.07\times)$ & $(1.00\times)$ \\
\bottomrule
\end{tabular}
}
\end{table}

\rebuttal{The ablation study in Table \ref{tab:burgers_64_ablation} examines how different warm start epochs influence the performance of PICore on FNO and UNO. We fix the coreset selection algorithm to EL2N and the dataset to Burgers. Across warm start configurations, extending the number of epochs beyond 10 to 25 or 50 epochs leads to only marginal gains, and by 100 epochs, the improvements are negligible or even slightly inconsistent within the bounds of standard deviation. For FNO, the 25-epoch variant achieves the lowest errors at lower data fractions (20–40\%), suggesting that moderate warm starting may yield slightly better initialization for PICore. UNO, on the other hand, shows stable performance across all pretraining lengths, implying that it benefits less from extended warm starting. Overall, the table suggests that minimal warm starting (around 10–25 epochs) is sufficient for both operator families, while additional finetuning offers little benefit.}

\subsection{Multi-Resolution Coreset Selection}
\begin{table}[H]
\centering
\caption{Burgers NRMSE with Multi Resolution Data}
\label{tab:burgers_multi_resolution}
\resizebox{\textwidth}{!}{%
\begin{tabular}{lllcccccc}
\toprule
Operator & Method & Algorithm & 20.0\% & 30.0\% & 40.0\% & 60.0\% & 80.0\% & 100.0\% \\
\midrule
\multirow{14}{*}{FNO} & \multirow{6}{*}{Supervised} & \texttt{craig} & $8.29 \pm 1.58 \times 10^{-2}$ & $7.73 \pm 0.83 \times 10^{-2}$ & $7.56 \pm 0.75 \times 10^{-2}$ & $7.78 \pm 0.88 \times 10^{-2}$ & $7.25 \pm 0.84 \times 10^{-2}$ & $6.85 \pm 0.79 \times 10^{-2}$ \\
 &  & \texttt{gradmatch} & $8.33 \pm 1.16 \times 10^{-2}$ & $8.10 \pm 0.93 \times 10^{-2}$ & $7.92 \pm 0.44 \times 10^{-2}$ & $7.64 \pm 0.81 \times 10^{-2}$ & $6.94 \pm 0.58 \times 10^{-2}$ & $6.85 \pm 0.79 \times 10^{-2}$ \\
 &  & \texttt{adacore} & $9.24 \pm 0.83 \times 10^{-2}$ & $8.73 \pm 0.54 \times 10^{-2}$ & $8.03 \pm 0.98 \times 10^{-2}$ & $7.44 \pm 0.93 \times 10^{-2}$ & $6.89 \pm 0.75 \times 10^{-2}$ & $6.85 \pm 0.79 \times 10^{-2}$ \\
 &  & \texttt{el2n} & $7.61 \pm 0.96 \times 10^{-2}$ & $\mathbf{7.34 \pm 1.30 \times 10^{-2}}$ & $7.55 \pm 0.60 \times 10^{-2}$ & $7.30 \pm 0.61 \times 10^{-2}$ & $6.94 \pm 0.85 \times 10^{-2}$ & $6.85 \pm 0.79 \times 10^{-2}$ \\
 &  & \texttt{graNd} & $8.44 \pm 0.73 \times 10^{-2}$ & $7.95 \pm 1.00 \times 10^{-2}$ & $7.75 \pm 0.88 \times 10^{-2}$ & $7.45 \pm 0.60 \times 10^{-2}$ & $7.23 \pm 0.58 \times 10^{-2}$ & $6.85 \pm 0.79 \times 10^{-2}$ \\
\cmidrule(lr){3-9}
 &  & \textit{Acceleration} & $3.75 \pm 0.00\times$ & $2.76 \pm 0.00\times$ & $2.24 \pm 0.00\times$ & $1.60 \pm 0.00\times$ & $1.24 \pm 0.00\times$ & $1.00 \pm 0.00\times$ \\
\cmidrule(lr){3-9}\morecmidrules\cmidrule(lr){3-9}
 & \multirow{6}{*}{PICore} & \texttt{craig} & $\mathbf{7.37 \pm 1.63 \times 10^{-2}}$ & $8.09 \pm 0.62 \times 10^{-2}$ & $7.70 \pm 1.20 \times 10^{-2}$ & $7.36 \pm 0.70 \times 10^{-2}$ & $7.31 \pm 0.45 \times 10^{-2}$ & $6.85 \pm 0.79 \times 10^{-2}$\\
 &  & \texttt{gradmatch} & $7.90 \pm 1.89 \times 10^{-2}$ & $8.07 \pm 1.55 \times 10^{-2}$ & $7.79 \pm 1.19 \times 10^{-2}$ & $7.54 \pm 0.59 \times 10^{-2}$ & $7.15 \pm 0.85 \times 10^{-2}$ & $6.85 \pm 0.79 \times 10^{-2}$\\
 &  & \texttt{adacore} & $8.56 \pm 1.78 \times 10^{-2}$ & $8.14 \pm 1.36 \times 10^{-2}$ & $8.20 \pm 0.81 \times 10^{-2}$ & $7.90 \pm 1.03 \times 10^{-2}$ & $7.34 \pm 1.10 \times 10^{-2}$ & $6.85 \pm 0.79 \times 10^{-2}$\\
 &  & \texttt{el2n} & $8.10 \pm 0.66 \times 10^{-2}$ & $7.37 \pm 1.87 \times 10^{-2}$ & $\mathbf{7.49 \pm 1.54 \times 10^{-2}}$ & $\mathbf{7.29 \pm 0.65 \times 10^{-2}}$ & $\mathbf{6.78 \pm 0.73 \times 10^{-2}}$ & $6.85 \pm 0.79 \times 10^{-2}$\\
 &  & \texttt{graNd} & $8.88 \pm 0.63 \times 10^{-2}$ & $8.39 \pm 0.87 \times 10^{-2}$ & $8.19 \pm 1.16 \times 10^{-2}$ & $7.56 \pm 0.61 \times 10^{-2}$ & $7.31 \pm 0.81 \times 10^{-2}$ & $6.85 \pm 0.79 \times 10^{-2}$\\
\cmidrule(lr){3-9}
 &  & \textit{Acceleration} & $5.08 \pm 0.01\times$ & $3.32 \pm 0.00\times$ & $2.54 \pm 0.00\times$ & $1.70 \pm 0.00\times$ & $1.27 \pm 0.00\times$ & $1.00 \pm 0.00\times$\\
\bottomrule
\end{tabular}
}
\end{table}

\rebuttal{Table \ref{tab:burgers_multi_resolution} compares Supervised and Physics-Informed Coreset Selection (PICore) across varying data resolutions for the Burgers’ equation using the FNO operator. In this experiment, the dataset is split into two equal parts: one half is generated at a resolution of 32, and the other at a resolution of 64, introducing multi-scale variability in the training data. Overall, PICore demonstrates improved stability and often superior accuracy at lower data percentages, particularly in the FNO setting. For instance, under FNO with 20–60\% data, PICore achieves consistently lower NRMSE values than supervised methods, especially with the \texttt{el2n} and \texttt{craig} algorithms, indicating better generalization when fewer samples are available. This advantage reflects the integration of physical constraints, which act as a strong inductive bias that helps retain key solution structures even when data is sparse or noisy. The mixed-resolution setup further amplifies this effect, as while supervised coresets can struggle to reconcile the differences between coarse (32) and fine (64) grids, PICore leverages physics-informed consistency to align features across scales, resulting in more stable and resolution-invariant representations of the Burgers’ dynamics. However, this setup also introduces greater variability and noise, as discrepancies between resolutions can create inconsistencies in gradient magnitudes and feature smoothness, but these issues are generally inherent to multi-resolution datasets, rather than specific to the coreset selection approach itself.}

\subsection{Zero Shot Super Resolution}
\label{subsec: zero shot super resolution}
Since Neural Operators learn parameters independently of the discretization (unlike PINNs), trained neural operators can perform zero-shot super-resolution, which allows for training a model at a lower resolution and evaluating at a higher resolution. We scale the Advection and Burgers datasets to resolutions of 128 and 256, the Darcy dataset to 128, and the Navier Stokes Incompressible dataset to size 256 in tables ~\ref{tab:advection_128_combined} –\ref{tab:navierstokesincompressible_256_combined}.

% The results for Advection and Burgers on the zero shot super resolution experiments are similar to the results at resolution 64, with the test NRMSE being slightly larger due to the increased resolution. However, one interesting observation is that 

\begin{table}[htbp]
\centering
\resizebox{0.8\textwidth}{!}{%
\begin{tabular}{lllcccccc}
\toprule
Operator & Method & Algorithm & 20.0\% & 30.0\% & 40.0\% & 60.0\% & 80.0\% & 100.0\% \\
\midrule
\multirow{12}{*}{FNO} & \multirow{6}{*}{Supervised} & \texttt{craig} & \textbf{$7.69 \times 10^{-2}$} & $8.41 \times 10^{-2}$ & $8.05 \times 10^{-2}$ & $8.04 \times 10^{-2}$ & $7.61 \times 10^{-2}$ & \textbf{$5.90 \times 10^{-2}$} \\
 &  & \texttt{gradmatch} & $8.38 \times 10^{-2}$ & $7.87 \times 10^{-2}$ & $7.85 \times 10^{-2}$ & $7.86 \times 10^{-2}$ & $7.73 \times 10^{-2}$ & \textbf{$5.90 \times 10^{-2}$} \\
 &  & \texttt{adacore} & $8.45 \times 10^{-2}$ & $7.85 \times 10^{-2}$ & $7.99 \times 10^{-2}$ & \textbf{$7.42 \times 10^{-2}$} & $7.43 \times 10^{-2}$ & \textbf{$5.90 \times 10^{-2}$} \\
 &  & \texttt{el2n} & $8.34 \times 10^{-2}$ & $8.05 \times 10^{-2}$ & $8.10 \times 10^{-2}$ & $7.88 \times 10^{-2}$ & $7.67 \times 10^{-2}$ & \textbf{$5.90 \times 10^{-2}$} \\
 &  & \texttt{graNd} & $8.25 \times 10^{-2}$ & $8.05 \times 10^{-2}$ & \textbf{$7.79 \times 10^{-2}$} & $7.60 \times 10^{-2}$ & \textbf{$7.26 \times 10^{-2}$} & \textbf{$5.90 \times 10^{-2}$} \\
\cmidrule(lr){3-9}
 & \multirow{6}{*}{PICore} & \texttt{craig} & $8.38 \times 10^{-2}$ & $8.01 \times 10^{-2}$ & $7.76 \times 10^{-2}$ & $7.41 \times 10^{-2}$ & $7.35 \times 10^{-2}$ & \textbf{$5.90 \times 10^{-2}$}\\
 &  & \texttt{gradmatch} & $8.85 \times 10^{-2}$ & $8.35 \times 10^{-2}$ & $8.15 \times 10^{-2}$ & $7.76 \times 10^{-2}$ & $7.59 \times 10^{-2}$ & \textbf{$5.90 \times 10^{-2}$}\\
 &  & \texttt{adacore} & $8.82 \times 10^{-2}$ & \textbf{$7.57 \times 10^{-2}$} & \textbf{$7.28 \times 10^{-2}$} & \textbf{$7.27 \times 10^{-2}$} & \textbf{$7.03 \times 10^{-2}$} & \textbf{$5.90 \times 10^{-2}$}\\
 &  & \texttt{el2n} & $8.95 \times 10^{-2}$ & $8.42 \times 10^{-2}$ & $8.33 \times 10^{-2}$ & $8.18 \times 10^{-2}$ & $7.98 \times 10^{-2}$ & \textbf{$5.90 \times 10^{-2}$}\\
 &  & \texttt{graNd} & $8.38 \times 10^{-2}$ & $8.15 \times 10^{-2}$ & $7.86 \times 10^{-2}$ & $7.72 \times 10^{-2}$ & $7.50 \times 10^{-2}$ & \textbf{$5.90 \times 10^{-2}$}\\
\midrule
\multirow{12}{*}{UNO} & \multirow{6}{*}{Supervised} & \texttt{craig} & $1.72 \times 10^{-1}$ & $1.66 \times 10^{-1}$ & $1.61 \times 10^{-1}$ & $1.48 \times 10^{-1}$ & \textbf{$1.14 \times 10^{-1}$} & \textbf{$1.09 \times 10^{-1}$} \\
 &  & \texttt{gradmatch} & \textbf{$1.70 \times 10^{-1}$} & \textbf{$1.65 \times 10^{-1}$} & \textbf{$1.61 \times 10^{-1}$} & \textbf{$1.42 \times 10^{-1}$} & $1.21 \times 10^{-1}$ & \textbf{$1.09 \times 10^{-1}$} \\
 &  & \texttt{adacore} & $2.25 \times 10^{-1}$ & $1.85 \times 10^{-1}$ & $1.71 \times 10^{-1}$ & $1.58 \times 10^{-1}$ & $1.27 \times 10^{-1}$ & \textbf{$1.09 \times 10^{-1}$} \\
 &  & \texttt{el2n} & $1.69 \times 10^{-1}$ & \textbf{$1.65 \times 10^{-1}$} & $1.62 \times 10^{-1}$ & $1.55 \times 10^{-1}$ & $1.25 \times 10^{-1}$ & \textbf{$1.09 \times 10^{-1}$} \\
 &  & \texttt{graNd} & $1.69 \times 10^{-1}$ & $1.64 \times 10^{-1}$ & \textbf{$1.61 \times 10^{-1}$} & $1.49 \times 10^{-1}$ & $1.24 \times 10^{-1}$ & \textbf{$1.09 \times 10^{-1}$} \\
\cmidrule(lr){3-9}
 & \multirow{6}{*}{PICore} & \texttt{craig} & $1.74 \times 10^{-1}$ & \textbf{$1.61 \times 10^{-1}$} & \textbf{$1.60 \times 10^{-1}$} & $1.51 \times 10^{-1}$ & $1.26 \times 10^{-1}$ & \textbf{$1.09 \times 10^{-1}$}\\
 &  & \texttt{gradmatch} & \textbf{$1.70 \times 10^{-1}$} & \textbf{$1.65 \times 10^{-1}$} & $1.62 \times 10^{-1}$ & $1.49 \times 10^{-1}$ & $1.23 \times 10^{-1}$ & \textbf{$1.09 \times 10^{-1}$}\\
 &  & \texttt{adacore} & $2.32 \times 10^{-1}$ & $1.82 \times 10^{-1}$ & $1.61 \times 10^{-1}$ & $1.50 \times 10^{-1}$ & $1.45 \times 10^{-1}$ & \textbf{$1.09 \times 10^{-1}$}\\
 &  & \texttt{el2n} & $1.69 \times 10^{-1}$ & $1.64 \times 10^{-1}$ & $1.63 \times 10^{-1}$ & $1.52 \times 10^{-1}$ & $1.27 \times 10^{-1}$ & \textbf{$1.09 \times 10^{-1}$}\\
 &  & \texttt{graNd} & \textbf{$1.70 \times 10^{-1}$} & $1.64 \times 10^{-1}$ & $1.62 \times 10^{-1}$ & \textbf{$1.53 \times 10^{-1}$} & \textbf{$1.23 \times 10^{-1}$} & \textbf{$1.09 \times 10^{-1}$}\\
\bottomrule
\end{tabular}
}
\caption{Test NRMSE on the Advection dataset at resolution 128 across varying coreset percentages (20\%–100\%) between supervised and PICore-based coreset selection methods using both FNO and UNO architectures.}
\label{tab:advection_128_combined}
\end{table}
\begin{table}[htbp]
\centering
\resizebox{0.8\textwidth}{!}{%
\begin{tabular}{lllcccccc}
\toprule
Operator & Method & Algorithm & 20.0\% & 30.0\% & 40.0\% & 60.0\% & 80.0\% & 100.0\% \\
\midrule
\multirow{12}{*}{FNO} & \multirow{6}{*}{Supervised} & \texttt{craig} & $8.55 \times 10^{-2}$ & $9.30 \times 10^{-2}$ & $8.99 \times 10^{-2}$ & $9.00 \times 10^{-2}$ & $8.64 \times 10^{-2}$ & $7.07 \times 10^{-2}$ \\
 &  & \texttt{gradmatch} & $9.22 \times 10^{-2}$ & $8.79 \times 10^{-2}$ & $8.80 \times 10^{-2}$ & $8.85 \times 10^{-2}$ & $8.76 \times 10^{-2}$ & $7.07 \times 10^{-2}$ \\
 &  & \texttt{adacore} & $9.07 \times 10^{-2}$ & $8.65 \times 10^{-2}$ & $8.87 \times 10^{-2}$ & $8.43 \times 10^{-2}$ & $8.47 \times 10^{-2}$ & $7.07 \times 10^{-2}$ \\
 &  & \texttt{el2n} & $9.18 \times 10^{-2}$ & $8.98 \times 10^{-2}$ & $9.04 \times 10^{-2}$ & $8.85 \times 10^{-2}$ & $8.67 \times 10^{-2}$ & $7.07 \times 10^{-2}$ \\
 &  & \texttt{graNd} & $9.09 \times 10^{-2}$ & $8.95 \times 10^{-2}$ & $8.76 \times 10^{-2}$ & $8.59 \times 10^{-2}$ & $8.31 \times 10^{-2}$ & $7.07 \times 10^{-2}$ \\
\cmidrule(lr){3-9}
 & \multirow{6}{*}{PICore} & \texttt{craig} & $9.15 \times 10^{-2}$ & $8.86 \times 10^{-2}$ & $8.67 \times 10^{-2}$ & $8.39 \times 10^{-2}$ & $8.38 \times 10^{-2}$ & $7.07 \times 10^{-2}$\\
 &  & \texttt{gradmatch} & $9.68 \times 10^{-2}$ & $9.28 \times 10^{-2}$ & $9.12 \times 10^{-2}$ & $8.79 \times 10^{-2}$ & $8.63 \times 10^{-2}$ & $7.07 \times 10^{-2}$\\
 &  & \texttt{adacore} & $9.40 \times 10^{-2}$ & $8.37 \times 10^{-2}$ & $8.20 \times 10^{-2}$ & $8.27 \times 10^{-2}$ & $8.08 \times 10^{-2}$ & $7.07 \times 10^{-2}$\\
 &  & \texttt{el2n} & $9.77 \times 10^{-2}$ & $9.30 \times 10^{-2}$ & $9.24 \times 10^{-2}$ & $9.13 \times 10^{-2}$ & $8.96 \times 10^{-2}$ & $7.07 \times 10^{-2}$\\
 &  & \texttt{graNd} & $9.23 \times 10^{-2}$ & $9.06 \times 10^{-2}$ & $8.84 \times 10^{-2}$ & $8.77 \times 10^{-2}$ & $8.56 \times 10^{-2}$ & $7.07 \times 10^{-2}$\\
\midrule
\multirow{12}{*}{UNO} & \multirow{6}{*}{Supervised} & \texttt{craig} & $1.86 \times 10^{-1}$ & $1.80 \times 10^{-1}$ & $1.76 \times 10^{-1}$ & $1.63 \times 10^{-1}$ & $1.35 \times 10^{-1}$ & $1.31 \times 10^{-1}$ \\
 &  & \texttt{gradmatch} & $1.84 \times 10^{-1}$ & $1.79 \times 10^{-1}$ & $1.77 \times 10^{-1}$ & $1.59 \times 10^{-1}$ & $1.41 \times 10^{-1}$ & $1.31 \times 10^{-1}$ \\
 &  & \texttt{adacore} & $2.30 \times 10^{-1}$ & $1.95 \times 10^{-1}$ & $1.83 \times 10^{-1}$ & $1.73 \times 10^{-1}$ & $1.46 \times 10^{-1}$ & $1.31 \times 10^{-1}$ \\
 &  & \texttt{el2n} & $1.83 \times 10^{-1}$ & $1.80 \times 10^{-1}$ & $1.78 \times 10^{-1}$ & $1.71 \times 10^{-1}$ & $1.44 \times 10^{-1}$ & $1.31 \times 10^{-1}$ \\
 &  & \texttt{graNd} & $1.82 \times 10^{-1}$ & $1.79 \times 10^{-1}$ & $1.76 \times 10^{-1}$ & $1.64 \times 10^{-1}$ & $1.43 \times 10^{-1}$ & $1.31 \times 10^{-1}$ \\
\cmidrule(lr){3-9}
 & \multirow{6}{*}{PICore} & \texttt{craig} & $1.87 \times 10^{-1}$ & $1.74 \times 10^{-1}$ & $1.75 \times 10^{-1}$ & $1.67 \times 10^{-1}$ & $1.45 \times 10^{-1}$ & $1.31 \times 10^{-1}$\\
 &  & \texttt{gradmatch} & $1.83 \times 10^{-1}$ & $1.80 \times 10^{-1}$ & $1.78 \times 10^{-1}$ & $1.65 \times 10^{-1}$ & $1.42 \times 10^{-1}$ & $1.31 \times 10^{-1}$\\
 &  & \texttt{adacore} & $2.38 \times 10^{-1}$ & $1.92 \times 10^{-1}$ & $1.75 \times 10^{-1}$ & $1.66 \times 10^{-1}$ & $1.61 \times 10^{-1}$ & $1.31 \times 10^{-1}$\\
 &  & \texttt{el2n} & $1.84 \times 10^{-1}$ & $1.79 \times 10^{-1}$ & $1.78 \times 10^{-1}$ & $1.68 \times 10^{-1}$ & $1.47 \times 10^{-1}$ & $1.31 \times 10^{-1}$\\
 &  & \texttt{graNd} & $1.83 \times 10^{-1}$ & $1.79 \times 10^{-1}$ & $1.77 \times 10^{-1}$ & $1.68 \times 10^{-1}$ & $1.43 \times 10^{-1}$ & $1.31 \times 10^{-1}$\\
\bottomrule
\end{tabular}
}
\caption{Test NRMSE on the Advection dataset at resolution 256 across varying coreset percentages (20\%–100\%) between supervised and PICore-based coreset selection methods using both FNO and UNO architectures.}
\label{tab:advection_256_combined}
\end{table}

\begin{table}[htbp]
\centering
\resizebox{0.8\textwidth}{!}{%
\begin{tabular}{lllcccccc}
\toprule
Operator & Method & Algorithm & 20.0\% & 30.0\% & 40.0\% & 60.0\% & 80.0\% & 100.0\% \\
\midrule
\multirow{12}{*}{FNO} & \multirow{6}{*}{Supervised} & \texttt{craig} & $6.83 \times 10^{-2}$ & $7.32 \times 10^{-2}$ & $7.20 \times 10^{-2}$ & $6.90 \times 10^{-2}$ & $6.14 \times 10^{-2}$ & $4.57 \times 10^{-2}$ \\
 &  & \texttt{gradmatch} & $6.56 \times 10^{-2}$ & $6.35 \times 10^{-2}$ & $6.68 \times 10^{-2}$ & $6.32 \times 10^{-2}$ & $6.36 \times 10^{-2}$ & $4.57 \times 10^{-2}$ \\
 &  & \texttt{adacore} & $6.15 \times 10^{-2}$ & $6.07 \times 10^{-2}$ & $6.72 \times 10^{-2}$ & $6.34 \times 10^{-2}$ & $6.28 \times 10^{-2}$ & $4.57 \times 10^{-2}$ \\
 &  & \texttt{el2n} & $6.86 \times 10^{-2}$ & $7.17 \times 10^{-2}$ & $6.74 \times 10^{-2}$ & $6.43 \times 10^{-2}$ & $6.08 \times 10^{-2}$ & $4.57 \times 10^{-2}$ \\
 &  & \texttt{graNd} & $6.72 \times 10^{-2}$ & $6.00 \times 10^{-2}$ & $6.47 \times 10^{-2}$ & $6.27 \times 10^{-2}$ & $6.05 \times 10^{-2}$ & $4.57 \times 10^{-2}$ \\
\cmidrule(lr){3-9}
 & \multirow{6}{*}{PICore} & \texttt{craig} & $6.67 \times 10^{-2}$ & $6.31 \times 10^{-2}$ & $6.38 \times 10^{-2}$ & $6.22 \times 10^{-2}$ & $6.39 \times 10^{-2}$ & $4.57 \times 10^{-2}$\\
 &  & \texttt{gradmatch} & $6.40 \times 10^{-2}$ & $6.62 \times 10^{-2}$ & $6.81 \times 10^{-2}$ & $6.47 \times 10^{-2}$ & $6.15 \times 10^{-2}$ & $4.57 \times 10^{-2}$\\
 &  & \texttt{adacore} & $7.39 \times 10^{-2}$ & $6.60 \times 10^{-2}$ & $6.59 \times 10^{-2}$ & $6.38 \times 10^{-2}$ & $6.46 \times 10^{-2}$ & $4.57 \times 10^{-2}$\\
 &  & \texttt{el2n} & $6.29 \times 10^{-2}$ & $6.36 \times 10^{-2}$ & $6.60 \times 10^{-2}$ & $6.29 \times 10^{-2}$ & $6.38 \times 10^{-2}$ & $4.57 \times 10^{-2}$\\
 &  & \texttt{graNd} & $6.71 \times 10^{-2}$ & $6.58 \times 10^{-2}$ & $6.91 \times 10^{-2}$ & $6.50 \times 10^{-2}$ & $6.45 \times 10^{-2}$ & $4.57 \times 10^{-2}$\\
\midrule
\multirow{12}{*}{UNO} & \multirow{6}{*}{Supervised} & \texttt{craig} & $2.94 \times 10^{-2}$ & $2.77 \times 10^{-2}$ & $2.62 \times 10^{-2}$ & $2.33 \times 10^{-2}$ & $2.19 \times 10^{-2}$ & $2.16 \times 10^{-2}$ \\
 &  & \texttt{gradmatch} & $3.00 \times 10^{-2}$ & $2.83 \times 10^{-2}$ & $2.67 \times 10^{-2}$ & $2.45 \times 10^{-2}$ & $2.31 \times 10^{-2}$ & $2.16 \times 10^{-2}$ \\
 &  & \texttt{adacore} & $4.36 \times 10^{-2}$ & $3.39 \times 10^{-2}$ & $2.98 \times 10^{-2}$ & $2.47 \times 10^{-2}$ & $2.23 \times 10^{-2}$ & $2.16 \times 10^{-2}$ \\
 &  & \texttt{el2n} & $2.94 \times 10^{-2}$ & $2.76 \times 10^{-2}$ & $2.62 \times 10^{-2}$ & $2.39 \times 10^{-2}$ & $2.28 \times 10^{-2}$ & $2.16 \times 10^{-2}$ \\
 &  & \texttt{graNd} & $3.01 \times 10^{-2}$ & $2.83 \times 10^{-2}$ & $2.69 \times 10^{-2}$ & $2.47 \times 10^{-2}$ & $2.30 \times 10^{-2}$ & $2.16 \times 10^{-2}$ \\
\cmidrule(lr){3-9}
 & \multirow{6}{*}{PICore} & \texttt{craig} & $3.18 \times 10^{-2}$ & $2.96 \times 10^{-2}$ & $2.77 \times 10^{-2}$ & $2.38 \times 10^{-2}$ & $2.15 \times 10^{-2}$ & $2.16 \times 10^{-2}$\\
 &  & \texttt{gradmatch} & $2.97 \times 10^{-2}$ & $2.76 \times 10^{-2}$ & $2.69 \times 10^{-2}$ & $2.40 \times 10^{-2}$ & $2.27 \times 10^{-2}$ & $2.16 \times 10^{-2}$\\
 &  & \texttt{adacore} & $4.77 \times 10^{-2}$ & $3.68 \times 10^{-2}$ & $3.09 \times 10^{-2}$ & $2.49 \times 10^{-2}$ & $2.19 \times 10^{-2}$ & $2.16 \times 10^{-2}$\\
 &  & \texttt{el2n} & $2.96 \times 10^{-2}$ & $2.77 \times 10^{-2}$ & $2.62 \times 10^{-2}$ & $2.43 \times 10^{-2}$ & $2.29 \times 10^{-2}$ & $2.16 \times 10^{-2}$\\
 &  & \texttt{graNd} & $2.89 \times 10^{-2}$ & $2.72 \times 10^{-2}$ & $2.66 \times 10^{-2}$ & $2.49 \times 10^{-2}$ & $2.29 \times 10^{-2}$ & $2.16 \times 10^{-2}$\\
\bottomrule
\end{tabular}
}
\caption{Test NRMSE on the Burgers dataset at resolution 128 across varying coreset percentages (20\%–100\%) between supervised and PICore-based coreset selection methods using both FNO and UNO architectures.}
\label{tab:burgers_128_combined}

\end{table}

\begin{table}[htbp]
\centering
\label{tab:burgers_256_combined}
\resizebox{0.8\textwidth}{!}{%
\begin{tabular}{lllcccccc}
\toprule
Operator & Method & Algorithm & 20.0\% & 30.0\% & 40.0\% & 60.0\% & 80.0\% & 100.0\% \\
\midrule
\multirow{12}{*}{FNO} & \multirow{6}{*}{Supervised} & \texttt{craig} & $7.02 \times 10^{-2}$ & $7.52 \times 10^{-2}$ & $7.40 \times 10^{-2}$ & $7.12 \times 10^{-2}$ & $6.37 \times 10^{-2}$ & $4.89 \times 10^{-2}$ \\
 &  & \texttt{gradmatch} & $6.72 \times 10^{-2}$ & $6.55 \times 10^{-2}$ & $6.89 \times 10^{-2}$ & $6.55 \times 10^{-2}$ & $6.60 \times 10^{-2}$ & $4.89 \times 10^{-2}$ \\
 &  & \texttt{adacore} & $6.26 \times 10^{-2}$ & $6.18 \times 10^{-2}$ & $6.90 \times 10^{-2}$ & $6.55 \times 10^{-2}$ & $6.52 \times 10^{-2}$ & $4.89 \times 10^{-2}$ \\
 &  & \texttt{el2n} & $7.04 \times 10^{-2}$ & $7.38 \times 10^{-2}$ & $6.96 \times 10^{-2}$ & $6.66 \times 10^{-2}$ & $6.32 \times 10^{-2}$ & $4.89 \times 10^{-2}$ \\
 &  & \texttt{graNd} & $6.89 \times 10^{-2}$ & $6.21 \times 10^{-2}$ & $6.69 \times 10^{-2}$ & $6.50 \times 10^{-2}$ & $6.29 \times 10^{-2}$ & $4.89 \times 10^{-2}$ \\
\cmidrule(lr){3-9}
 & \multirow{6}{*}{PICore} & \texttt{craig} & $6.85 \times 10^{-2}$ & $6.52 \times 10^{-2}$ & $6.59 \times 10^{-2}$ & $6.45 \times 10^{-2}$ & $6.63 \times 10^{-2}$ & $4.89 \times 10^{-2}$\\
 &  & \texttt{gradmatch} & $6.57 \times 10^{-2}$ & $6.83 \times 10^{-2}$ & $7.02 \times 10^{-2}$ & $6.70 \times 10^{-2}$ & $6.39 \times 10^{-2}$ & $4.89 \times 10^{-2}$\\
 &  & \texttt{adacore} & $7.50 \times 10^{-2}$ & $6.75 \times 10^{-2}$ & $6.80 \times 10^{-2}$ & $6.61 \times 10^{-2}$ & $6.69 \times 10^{-2}$ & $4.89 \times 10^{-2}$\\
 &  & \texttt{el2n} & $6.49 \times 10^{-2}$ & $6.57 \times 10^{-2}$ & $6.82 \times 10^{-2}$ & $6.52 \times 10^{-2}$ & $6.62 \times 10^{-2}$ & $4.89 \times 10^{-2}$\\
 &  & \texttt{graNd} & $6.88 \times 10^{-2}$ & $6.78 \times 10^{-2}$ & $7.12 \times 10^{-2}$ & $6.72 \times 10^{-2}$ & $6.69 \times 10^{-2}$ & $4.89 \times 10^{-2}$\\
\midrule
\multirow{12}{*}{UNO} & \multirow{6}{*}{Supervised} & \texttt{craig} & $3.38 \times 10^{-2}$ & $3.36 \times 10^{-2}$ & $3.26 \times 10^{-2}$ & $3.06 \times 10^{-2}$ & $2.96 \times 10^{-2}$ & $2.92 \times 10^{-2}$ \\
 &  & \texttt{gradmatch} & $3.52 \times 10^{-2}$ & $3.43 \times 10^{-2}$ & $3.36 \times 10^{-2}$ & $3.22 \times 10^{-2}$ & $3.10 \times 10^{-2}$ & $2.92 \times 10^{-2}$ \\
 &  & \texttt{adacore} & $4.70 \times 10^{-2}$ & $3.84 \times 10^{-2}$ & $3.51 \times 10^{-2}$ & $3.17 \times 10^{-2}$ & $3.01 \times 10^{-2}$ & $2.92 \times 10^{-2}$ \\
 &  & \texttt{el2n} & $3.46 \times 10^{-2}$ & $3.33 \times 10^{-2}$ & $3.27 \times 10^{-2}$ & $3.11 \times 10^{-2}$ & $3.05 \times 10^{-2}$ & $2.92 \times 10^{-2}$ \\
 &  & \texttt{graNd} & $3.51 \times 10^{-2}$ & $3.40 \times 10^{-2}$ & $3.34 \times 10^{-2}$ & $3.22 \times 10^{-2}$ & $3.11 \times 10^{-2}$ & $2.92 \times 10^{-2}$ \\
\cmidrule(lr){3-9}
 & \multirow{6}{*}{PICore} & \texttt{craig} & $3.63 \times 10^{-2}$ & $3.47 \times 10^{-2}$ & $3.36 \times 10^{-2}$ & $3.11 \times 10^{-2}$ & $2.97 \times 10^{-2}$ & $2.92 \times 10^{-2}$\\
 &  & \texttt{gradmatch} & $3.50 \times 10^{-2}$ & $3.35 \times 10^{-2}$ & $3.34 \times 10^{-2}$ & $3.14 \times 10^{-2}$ & $3.08 \times 10^{-2}$ & $2.92 \times 10^{-2}$\\
 &  & \texttt{adacore} & $5.08 \times 10^{-2}$ & $4.06 \times 10^{-2}$ & $3.57 \times 10^{-2}$ & $3.12 \times 10^{-2}$ & $2.95 \times 10^{-2}$ & $2.92 \times 10^{-2}$\\
 &  & \texttt{el2n} & $3.47 \times 10^{-2}$ & $3.36 \times 10^{-2}$ & $3.29 \times 10^{-2}$ & $3.16 \times 10^{-2}$ & $3.09 \times 10^{-2}$ & $2.92 \times 10^{-2}$\\
 &  & \texttt{graNd} & $3.41 \times 10^{-2}$ & $3.34 \times 10^{-2}$ & $3.34 \times 10^{-2}$ & $3.28 \times 10^{-2}$ & $3.12 \times 10^{-2}$ & $2.92 \times 10^{-2}$\\
\bottomrule
\end{tabular}
}
\caption{Test NRMSE on the Burgers dataset at resolution 256 across varying coreset percentages (20\%–100\%) between supervised and PICore-based coreset selection methods using both FNO and UNO architectures.}
\end{table}

\begin{table}[htbp]
\centering
\label{tab:darcy_128_combined}
\resizebox{0.8\textwidth}{!}{%
\begin{tabular}{lllcccccc}
\toprule
Operator & Method & Algorithm & 20.0\% & 30.0\% & 40.0\% & 60.0\% & 80.0\% & 100.0\% \\
\midrule
\multirow{12}{*}{FNO} & \multirow{6}{*}{Supervised} & \texttt{craig} & $1.67 \times 10^{-1}$ & $1.51 \times 10^{-1}$ & $1.36 \times 10^{-1}$ & $1.15 \times 10^{-1}$ & $1.02 \times 10^{-1}$ & $8.76 \times 10^{-2}$ \\
 &  & \texttt{gradmatch} & $1.72 \times 10^{-1}$ & $1.47 \times 10^{-1}$ & $1.36 \times 10^{-1}$ & $1.15 \times 10^{-1}$ & $1.04 \times 10^{-1}$ & $8.76 \times 10^{-2}$ \\
 &  & \texttt{adacore} & $1.88 \times 10^{-1}$ & $1.61 \times 10^{-1}$ & $1.41 \times 10^{-1}$ & $1.18 \times 10^{-1}$ & $1.05 \times 10^{-1}$ & $8.76 \times 10^{-2}$ \\
 &  & \texttt{el2n} & $1.68 \times 10^{-1}$ & $1.42 \times 10^{-1}$ & $1.33 \times 10^{-1}$ & $1.15 \times 10^{-1}$ & $1.02 \times 10^{-1}$ & $8.76 \times 10^{-2}$ \\
 &  & \texttt{graNd} & $1.67 \times 10^{-1}$ & $1.46 \times 10^{-1}$ & $1.32 \times 10^{-1}$ & $1.16 \times 10^{-1}$ & $1.04 \times 10^{-1}$ & $8.76 \times 10^{-2}$ \\
\cmidrule(lr){3-9}
 & \multirow{6}{*}{PICore} & \texttt{craig} & $1.68 \times 10^{-1}$ & $1.53 \times 10^{-1}$ & $1.39 \times 10^{-1}$ & $1.16 \times 10^{-1}$ & $1.03 \times 10^{-1}$ & $8.76 \times 10^{-2}$\\
 &  & \texttt{gradmatch} & $1.73 \times 10^{-1}$ & $1.48 \times 10^{-1}$ & $1.38 \times 10^{-1}$ & $1.16 \times 10^{-1}$ & $1.03 \times 10^{-1}$ & $8.76 \times 10^{-2}$\\
 &  & \texttt{adacore} & $1.98 \times 10^{-1}$ & $1.58 \times 10^{-1}$ & $1.37 \times 10^{-1}$ & $1.10 \times 10^{-1}$ & $1.02 \times 10^{-1}$ & $8.76 \times 10^{-2}$\\
 &  & \texttt{el2n} & $1.66 \times 10^{-1}$ & $1.48 \times 10^{-1}$ & $1.32 \times 10^{-1}$ & $1.13 \times 10^{-1}$ & $1.03 \times 10^{-1}$ & $8.76 \times 10^{-2}$\\
 &  & \texttt{graNd} & $1.72 \times 10^{-1}$ & $1.47 \times 10^{-1}$ & $1.35 \times 10^{-1}$ & $1.13 \times 10^{-1}$ & $1.03 \times 10^{-1}$ & $8.76 \times 10^{-2}$\\
\midrule
\multirow{12}{*}{UNO} & \multirow{6}{*}{Supervised} & \texttt{craig} & $1.36 \times 10^{0}$ & $1.57 \times 10^{0}$ & $1.57 \times 10^{0}$ & $1.48 \times 10^{0}$ & $1.40 \times 10^{0}$ & $1.72 \times 10^{0}$ \\
 &  & \texttt{gradmatch} & $1.65 \times 10^{0}$ & $1.57 \times 10^{0}$ & $1.69 \times 10^{0}$ & $1.74 \times 10^{0}$ & $1.55 \times 10^{0}$ & $1.72 \times 10^{0}$ \\
 &  & \texttt{adacore} & $1.31 \times 10^{0}$ & $1.29 \times 10^{0}$ & $1.39 \times 10^{0}$ & $1.38 \times 10^{0}$ & $1.40 \times 10^{0}$ & $1.72 \times 10^{0}$ \\
 &  & \texttt{el2n} & $1.61 \times 10^{0}$ & $1.53 \times 10^{0}$ & $1.66 \times 10^{0}$ & $1.53 \times 10^{0}$ & $1.58 \times 10^{0}$ & $1.72 \times 10^{0}$ \\
 &  & \texttt{graNd} & $1.75 \times 10^{0}$ & $1.59 \times 10^{0}$ & $1.57 \times 10^{0}$ & $1.48 \times 10^{0}$ & $1.52 \times 10^{0}$ & $1.72 \times 10^{0}$ \\
\cmidrule(lr){3-9}
 & \multirow{6}{*}{PICore} & \texttt{craig} & $1.37 \times 10^{0}$ & $1.55 \times 10^{0}$ & $1.51 \times 10^{0}$ & $1.59 \times 10^{0}$ & $1.51 \times 10^{0}$ & $1.72 \times 10^{0}$\\
 &  & \texttt{gradmatch} & $1.76 \times 10^{0}$ & $1.59 \times 10^{0}$ & $1.56 \times 10^{0}$ & $1.53 \times 10^{0}$ & $1.49 \times 10^{0}$ & $1.72 \times 10^{0}$\\
 &  & \texttt{adacore} & $1.22 \times 10^{0}$ & $1.29 \times 10^{0}$ & $1.25 \times 10^{0}$ & $1.30 \times 10^{0}$ & $1.54 \times 10^{0}$ & $1.72 \times 10^{0}$\\
 &  & \texttt{el2n} & $1.67 \times 10^{0}$ & $1.59 \times 10^{0}$ & $1.59 \times 10^{0}$ & $1.54 \times 10^{0}$ & $1.52 \times 10^{0}$ & $1.72 \times 10^{0}$\\
 &  & \texttt{graNd} & $1.77 \times 10^{0}$ & $1.57 \times 10^{0}$ & $1.59 \times 10^{0}$ & $1.55 \times 10^{0}$ & $1.52 \times 10^{0}$ & $1.72 \times 10^{0}$\\
\bottomrule
\end{tabular}
}
\caption{Test NRMSE on the Darcy dataset at resolution 128 across varying coreset percentages (20\%–100\%) between supervised and PICore-based coreset selection methods using both FNO and UNO architectures.}
\end{table}

\begin{table}[htbp]
\centering
\resizebox{0.8\textwidth}{!}{%
\begin{tabular}{lllcccccc}
\toprule
Operator & Method & Algorithm & 20.0\% & 30.0\% & 40.0\% & 60.0\% & 80.0\% & 100.0\% \\
\midrule
\multirow{12}{*}{FNO} & \multirow{6}{*}{Supervised} & \texttt{craig} & $2.13 \times 10^{-1}$ & $7.81 \times 10^{-2}$ & $7.24 \times 10^{-2}$ & $7.33 \times 10^{-2}$ & $7.05 \times 10^{-2}$ & $8.54 \times 10^{-2}$ \\
 &  & \texttt{gradmatch} & $1.57 \times 10^{-1}$ & $7.36 \times 10^{-2}$ & $7.24 \times 10^{-2}$ & $7.55 \times 10^{-2}$ & $7.49 \times 10^{-2}$ & $8.54 \times 10^{-2}$ \\
 &  & \texttt{adacore} & $3.18 \times 10^{-1}$ & $6.85 \times 10^{-2}$ & $6.16 \times 10^{-2}$ & $6.55 \times 10^{-2}$ & $6.48 \times 10^{-2}$ & $8.54 \times 10^{-2}$ \\
 &  & \texttt{el2n} & $1.43 \times 10^{-1}$ & $7.76 \times 10^{-2}$ & $7.38 \times 10^{-2}$ & $7.93 \times 10^{-2}$ & $7.64 \times 10^{-2}$ & $8.54 \times 10^{-2}$ \\
 &  & \texttt{graNd} & $1.67 \times 10^{-1}$ & $7.42 \times 10^{-2}$ & $6.78 \times 10^{-2}$ & $7.31 \times 10^{-2}$ & $6.98 \times 10^{-2}$ & $8.54 \times 10^{-2}$ \\
\cmidrule(lr){3-9}
 & \multirow{6}{*}{PICore} & \texttt{craig} & $1.73 \times 10^{-1}$ & $7.56 \times 10^{-2}$ & $6.50 \times 10^{-2}$ & $6.72 \times 10^{-2}$ & $6.75 \times 10^{-2}$ & $8.54 \times 10^{-2}$\\
 &  & \texttt{gradmatch} & $1.70 \times 10^{-1}$ & $7.83 \times 10^{-2}$ & $7.36 \times 10^{-2}$ & $7.61 \times 10^{-2}$ & $7.76 \times 10^{-2}$ & $8.54 \times 10^{-2}$\\
 &  & \texttt{adacore} & $2.84 \times 10^{-1}$ & $1.04 \times 10^{-1}$ & $6.38 \times 10^{-2}$ & $6.38 \times 10^{-2}$ & $6.90 \times 10^{-2}$ & $8.54 \times 10^{-2}$\\
 &  & \texttt{el2n} & $1.63 \times 10^{-1}$ & $7.51 \times 10^{-2}$ & $7.04 \times 10^{-2}$ & $7.61 \times 10^{-2}$ & $7.65 \times 10^{-2}$ & $8.54 \times 10^{-2}$\\
 &  & \texttt{graNd} & $1.55 \times 10^{-1}$ & $7.83 \times 10^{-2}$ & $7.57 \times 10^{-2}$ & $7.53 \times 10^{-2}$ & $7.75 \times 10^{-2}$ & $8.54 \times 10^{-2}$\\
\midrule
\multirow{12}{*}{UNO} & \multirow{6}{*}{Supervised} & \texttt{craig} & $5.99 \times 10^{-2}$ & $5.79 \times 10^{-2}$ & $5.64 \times 10^{-2}$ & $5.56 \times 10^{-2}$ & $5.52 \times 10^{-2}$ & $5.16 \times 10^{-2}$ \\
 &  & \texttt{gradmatch} & $5.99 \times 10^{-2}$ & $5.83 \times 10^{-2}$ & $5.68 \times 10^{-2}$ & $5.60 \times 10^{-2}$ & $5.53 \times 10^{-2}$ & $5.16 \times 10^{-2}$ \\
 &  & \texttt{adacore} & $6.45 \times 10^{-2}$ & $5.98 \times 10^{-2}$ & $5.77 \times 10^{-2}$ & $5.58 \times 10^{-2}$ & $5.49 \times 10^{-2}$ & $5.16 \times 10^{-2}$ \\
 &  & \texttt{el2n} & $5.93 \times 10^{-2}$ & $5.77 \times 10^{-2}$ & $5.66 \times 10^{-2}$ & $5.57 \times 10^{-2}$ & $5.52 \times 10^{-2}$ & $5.16 \times 10^{-2}$ \\
 &  & \texttt{graNd} & $5.94 \times 10^{-2}$ & $5.82 \times 10^{-2}$ & $5.69 \times 10^{-2}$ & $5.60 \times 10^{-2}$ & $5.52 \times 10^{-2}$ & $5.16 \times 10^{-2}$ \\
\cmidrule(lr){3-9}
 & \multirow{6}{*}{PICore} & \texttt{craig} & $6.07 \times 10^{-2}$ & $5.74 \times 10^{-2}$ & $5.66 \times 10^{-2}$ & $5.53 \times 10^{-2}$ & $5.51 \times 10^{-2}$ & $5.16 \times 10^{-2}$\\
 &  & \texttt{gradmatch} & $5.97 \times 10^{-2}$ & $5.76 \times 10^{-2}$ & $5.67 \times 10^{-2}$ & $5.60 \times 10^{-2}$ & $5.55 \times 10^{-2}$ & $5.16 \times 10^{-2}$\\
 &  & \texttt{adacore} & $6.56 \times 10^{-2}$ & $6.02 \times 10^{-2}$ & $5.81 \times 10^{-2}$ & $5.57 \times 10^{-2}$ & $5.50 \times 10^{-2}$ & $5.16 \times 10^{-2}$\\
 &  & \texttt{el2n} & $5.96 \times 10^{-2}$ & $5.76 \times 10^{-2}$ & $5.68 \times 10^{-2}$ & $5.58 \times 10^{-2}$ & $5.53 \times 10^{-2}$ & $5.16 \times 10^{-2}$\\
 &  & \texttt{graNd} & $5.96 \times 10^{-2}$ & $5.77 \times 10^{-2}$ & $5.65 \times 10^{-2}$ & $5.58 \times 10^{-2}$ & $5.54 \times 10^{-2}$ & $5.16 \times 10^{-2}$\\
\bottomrule
\end{tabular}
}
\caption{Test NRMSE on the Navier Stokes Incompressible dataset at resolution 256 across varying coreset percentages (20\%–100\%) between supervised and PICore-based coreset selection methods using both FNO and UNO architectures.}
\label{tab:navierstokesincompressible_256_combined}
\end{table}

\end{document}